\documentclass{article}
\usepackage{arxiv}
\usepackage[utf8]{inputenc} 
\usepackage[T1]{fontenc}    
\usepackage[hidelinks]{hyperref}       
\usepackage{booktabs}       
\usepackage{amsfonts}       
\usepackage{nicefrac}       
\usepackage{microtype}      
\usepackage{graphicx}
\usepackage{natbib}
 \bibpunct[, ]{(}{)}{,}{a}{}{,}%
\usepackage{doi}
\usepackage{tikz}
\usetikzlibrary{shapes.geometric, arrows}
\usepackage{caption}
\usepackage{rotating}
\usepackage{algorithm}
\usepackage{amsmath}
\usepackage{amssymb}
\DeclareMathOperator*{\argmax}{arg\,max}
\DeclareMathOperator*{\argmin}{arg\,min}
\usepackage[noend]{algpseudocode}
\usepackage{float}
\newfloat{algorithm}{t}{lop}
\usepackage{subfig} 
\usepackage{adjustbox} 
\usepackage{setspace}
\usepackage{pdflscape}
\usepackage{longtable}
\usepackage{threeparttable}
\usepackage{fancyhdr}
\title{Spatial-temporal-demand clustering for solving large-scale vehicle routing problems with time windows}
\date{}
\author{Christoph Kerscher$^{1}$, Stefan Minner$^{1,2}$\\
	$^{1}$Logistics and Supply Chain Management, School of Management, Technical University of Munich, Munich, Germany\\ 
 $^{2}$Munich Data Science Institute (MDSI), Technical University of Munich, Munich, Germany\\
}
\begin{document}
\centerline{\Large \bfseries Spatial-temporal-demand clustering for solving}
\vspace{0.1cm}
\centerline{\Large \bfseries large-scale vehicle routing problems with time windows}
\vspace{1cm}
\centerline{\large \bfseries Christoph Kerscher$^{1}$, Stefan Minner$^{1,2}$}
\vspace{0.1cm}
\centerline{$^{1}$Logistics and Supply Chain Management, School of Management, Technical University of Munich, Munich, Germany}
\centerline{$^{2}$Munich Data Science Institute (MDSI), Technical University of Munich, Munich, Germany}
\vspace{1cm}
\begin{abstract}
Several metaheuristics use decomposition and pruning strategies to solve large-scale instances of the vehicle routing problem (VRP). Those complexity reduction techniques often rely on simple, problem-specific rules. However, the growth in available data and advances in computer hardware enable data-based approaches that use machine learning (ML) to improve scalability of solution algorithms. We propose a decompose-route-improve (DRI) framework that groups customers using clustering. Its similarity metric incorporates customers' spatial, temporal, and demand data and is formulated to reflect the problem's objective function and constraints. The resulting sub-routing problems can independently be solved using any suitable algorithm. We apply pruned local search (LS) between solved subproblems to improve the overall solution. Pruning is based on customers' similarity information obtained in the decomposition phase. In a computational study, we parameterize and compare existing clustering algorithms and benchmark the DRI against the Hybrid Genetic Search (HGS) of \citet{Vidal_13_HGS-TW}. Results show that our data-based approach outperforms classic cluster-first, route-second approaches solely based on customers' spatial information. The newly introduced similarity metric forms separate sub-VRPs and improves the selection of LS moves in the improvement phase. Thus, the DRI scales existing metaheuristics to achieve high-quality solutions faster for large-scale VRPs by efficiently reducing complexity. Further, the DRI can be easily adapted to various solution methods and VRP characteristics, such as distribution of customer locations and demands, depot location, and different time window scenarios, making it a generalizable approach to solving routing problems.
\end{abstract}
\section{Introduction}
Given its practical relevance, numerous studies about the vehicle routing problem \citep[VRP,][]{Dantzig_1959_vrp} and its variants exist \citep{Vidal_20_vrp_variants}. Typically, rich VRPs of large-scale real-world applications are solved by heuristics.
State-of-the-art metaheuristics use most of their computation time searching for local improvements in an incumbent solution by modifying customer sequences within a given tour or changing customer-vehicle assignments \citep{Vidal_13_HGS-TW}. As the problem size increases, the number of possible local search (LS) operations grows exponentially. In response, complexity reduction techniques are applied to limit the solution space. Decomposition and aggregation methods divide the original problem into multiple smaller ones that are solved independently \citep{Santini_a_23_decomp_HGS}. Pruning limits the LS operators' exploration of new solutions \citep{Arnold_19_gls_pruning}. These strategies mostly follow simple rules tailored to a particular problem. In practice, however, solution algorithms must be scalable and adjustable to various problem characteristics. Thus, we propose a generalizable solution framework called decompose-route-improve (DRI) that reduces the complexity of large-scale routing problems using data-based decomposition. This approach uses unsupervised clustering to split the customers of a VRP into separate subsets. Its similarity metric combines customers' spatial, temporal, and demand features with the problem's objective function and constraints.
The resulting stand-alone small-sized sub-VRPs are solved independently. The solution to the overall problem is the combination of the individual solutions of the subproblems. Finally, LS, pruned based on customers' spatial-temporal-demand similarity, resolves unfavorable routing decisions at the perimeters of the subproblems.
This approach demonstrates high scalability and expeditiously achieves high-quality solutions for large-scale VRPs. 
Thus, our contribution to the growing research field of heuristic decomposition is twofold.
\newpage
\begin{itemize}
    \item[1.] We formulate a novel similarity metric for unsupervised clustering to improve scalability of state-of-the-art solution methods for large-scale routing problems.
    \item[2.] We tune hyperparameters of the DRI to reduce complexity based on problem attributes, i.e., size, customer characteristics, and fleet properties.
\end{itemize}
The remainder of this paper is structured as follows. Section \ref{lit} gives an overview of relevant literature. The model of the routing problem is formalized in Section \ref{prob_statement}. Section \ref{sol_frame} presents the DRI and data-based similarity metric. Section \ref{cs} describes the computational study and numerical results, including hyperparameter setups and a benchmark against a state-of-the-art metaheuristic. Section \ref{cnclsn} summarizes the key contributions and concludes the paper with future research directions.

\section{Literature review}\label{lit}
We present state-of-the-art solution methods and rule- and learning-based approaches to address the methods' scalability issues. At the end of this section, Table \ref{tab:lit_review} classifies the reviewed methods based on the problem variant they solve, applied scalability strategies, and the solution algorithm.
\subsection{State-of-the-art solution methods}
In the VRP with time windows (VRPTW) \citep{Braysy_05_VRPTW}, delivery time windows are added to customer requests. The arrival of a vehicle at a customer is only allowed within its specified time window. 
Recent branch-price-and-cut algorithms can solve VRPTW instances with up to 200 customers to optimality \citep[e.g., ][]{Costa_a_19_VRP_exact, Sahin_22_exact_VRPTW}. However, several hundreds or thousands of customer requests must be fulfilled in real-world distribution scenarios. The complexity of these large-scale problems makes exact solution methods impractical. Thus, state-of-the-art solution algorithms are metaheuristics that find feasible solutions of high quality in reasonable time. \citet{Arnold_19_v_large_scale} solve instances with up to $30,000$ customers of the capacitated VRP (CVRP). The very large-scale benchmark problems are based on real-world data of Belgium's daily parcel delivery services. Once an initial solution is found using the savings heuristic by \citet{Clarke_a_64_sav} and its routes are optimized using the 2opt-heuristic by \citet{Lin_a_73_2opt}, they apply heuristic pruning exclusively based on spatial distances between customers to limit the size of the neighborhoods explored by LS.
The Hybrid Genetic Search (HGS) \citep{Vidal_12_HGS_org} is the most effective solver for a wide variety of small- and mid-size routing problems, including the VRPTW \citep{Vidal_13_HGS-TW}. The HGS combines principles of genetic algorithms (GA), which create offspring solutions by crossing over its parent solutions, and inter- and intra-LS operations that try to improve the routing of that solution. Diversification and intensification of the search are achieved by an advanced population diversity management that accommodates feasible and infeasible solutions. 
\subsection{Scalability strategies}
\citet{Santini_a_23_decomp_HGS} demonstrate that splitting a routing problem into a set of subproblems using unsupervised clustering and solving them recursively by the HGS improves solution quality for CVRP instances with up to 1000 customers. Their decomposition strategy is route-based following the \textit{route-first, cluster-second (r-f, c-s)} principle and integrated repetitively, after a given number of iterations, into the HGS. Clustering is solely based on geographical distances between route centers.
\citet{Bent_a_10_temp_decomp_vrp} show that r-f, c-s decomposition based on customer information outperforms route-based separations for a large neighborhood search (LNS) heuristic. Subproblems are formed by customers within a randomly selected spatial (time) slice of the geographical plane (operational period). They report that decomposition based on customer features leads to better solutions faster. 
In r-f, c-s approaches, decomposition quality is limited by the incumbent route plan. This solution is found without any complexity reduction technique.
Even with the implementation of decomposition techniques, LS operations consume most of the computation time \citep[e.g.,][]{Ropke_a_06_ALNS, Vidal_13_HGS-TW, Accorsi_a_21_filo}. Heuristic pruning improves scalability of solution algorithms that rely considerably on creating solution neighborhoods. It limits LS operators in their exploration of new solutions. The overarching goal is to find high-quality solutions faster while using less memory. Limiting the size of neighborhoods by constraining LS moves among relatively close customers is among the most widely applied pruning techniques \citep[e.g.,][]{Helsgaun_00_effective, Toth_a_03_pruning, Arnold_19_v_large_scale}. However, these search limitation methodologies may exclude relevant edges, especially if routing decisions are not solely based on travel costs like in the VRPTW. Pruning that goes beyond strict rule-based restrictions is done in \citet{Beek_a_18_pruning_ls} and \citet{Arnold_a_21_PILS}. \citet{Beek_a_18_pruning_ls} limit the size of their in-memory stored Static Move Descriptors (SMDs) based on their contribution to the overall objective.\\\\ \citet{Arnold_a_21_PILS} exploit the information of extracted routing patterns of historical solutions on LS operators for the CVRP to reduce the heuristic's computational effort. None of the presented works combine decomposition and pruning in the solution method.
\subsection{Clustering approaches}
Unlike the decomposition strategies mentioned above, \textit{cluster-first, route-second (c-f, r-s)} approaches decompose a routing problem into less complex subproblems before the routing step. One of the first methods following this principle is the sweep algorithm \citep{Gillett_74_sweep}. It first assigns customers to a vehicle based on their polar angles and then does the sequencing of the individual routes. \citet{Fisher_81_gap_vrp} heuristically solve a generalized assignment problem to assign customers to vehicles. This results in multiple traveling salesman problems (TSPs) that are solved independently. \citet{Wong_84_del_area} define subareas of customers to be visited based on their historical requests. Customers in the same area are served by a single vehicle. This leads to intuitive routing decisions but often to an inefficient tour plan, especially if time windows are present, as the areas are fixed once determined. \citet{Schneider_15_tbra_tw} analyze the effect of time windows on the formation of customer districts. These territories are generated based on defined rules that take samples of the customers' spatial and temporal information and their historical demands into account. In a second step, a tabu search (TS) heuristic solves a TSP for every customer subset. All these works have in common that the decomposition step yields stand-alone TSP problems. Thus, the clustering restricts the routing decisions as the customer allocation to a vehicle is fixed. In the clustered vehicle routing problem (CluVRP), customers are pre-assigned to clusters, and a vehicle passes to more than one cluster on its route as the sum of the customers' demand per cluster is usually smaller than the vehicle capacity \citep{Battarra_14_cluvrp}.
In practical applications, however, such clusters are not known in advance. Unsupervised clustering allocates customers into multiple groups based on a specified similarity metric. \citet{Jain_10_clst_survey} provides a detailed summary of clustering methods most widely used, including $k$-means and its variants, hierarchical clustering algorithms like agglomerative clustering, and non-deterministic (e.g., fuzzy c-means) approaches that compute the degree of membership - i.e., the relative similarity - for every data object to all clusters. \citet{Yucenur_11_clst_ga_mdvrp} propose a genetic clustering algorithm to assign customers to depots in the multi-depot VRP (MDVRP). \citet{Ewbank_16_fuzzy_clst_cvrp} use a fuzzy clustering technique. The resulting TSPs are solved approximately using the nearest-neighbor (NN) and 2opt-heuristic. \citet{Qi_a_12_st_clst_vrp} solve large-scale VRPTWs with 1000 customers using $k$-medoid clustering and the I1 heuristic by \citet{Solomon_87_vrptw_bm}. Customers are grouped into subproblems based on their pairwise spatial distance and the temporal distance between their time windows. In all these works, clustering metrics are exclusively based on the distance between customers. In contrast, our clustering metric evaluates the similarity of customers concerning the contribution to the overall problem's objective and the consumption of the available resources when visited in sequence. Thus, our clustering is based on geographical distances between customers and their temporal and demand data.
\begin{table}[ht]
    \centering
    \begin{threeparttable}
        \caption{\label{tab:lit_review} Overview of contributions, applied scalability strategies, and solution algorithms}
        \begin{tabular}{ll|lll|ll|l}
    \toprule
    {} & {routing} & \multicolumn{3}{c|}{decomposition} & \multicolumn{2}{c|}{pruning} & {solution} \\ 
    {} & problem  & strategy  & feat.\tnote{1}  & subprob. & follows  & feat.\tnote{1} & algorithm\tnote{2} \\ 
    \midrule 
    \citet{Accorsi_a_21_filo} & CVRP & - & - & - & data & s & SA+ILS \\ 
    \citet{Arnold_19_v_large_scale} & CVRP & - & - & - & rules & s & LS \\ 
    \citet{Arnold_a_21_PILS} & CVRP & - & - & - & data & s & LS \\ 
    \citet{Beek_a_18_pruning_ls} & CVRP & - & - & - & data & s & LS \\ 
    \citet{Bent_a_10_temp_decomp_vrp} & VRPTW & r-f, c-s & st & VRPTW & - & - & LNS \\ 
    \citet{Ewbank_16_fuzzy_clst_cvrp} & CVRP & c-f, r-s & s & TSP & - & - & NN+2opt \\ 
    \citet{Gillett_74_sweep} & CVRP & c-f, r-s & s & TSP & - & - & 2opt \\ 
    \citet{Fisher_81_gap_vrp} & CVRP & c-f, r-s & s & TSP & - & - & B\&b \\ 
    \citet{Qi_a_12_st_clst_vrp} & VRPTW & c-f, r-s & st & VRPTW & - & - & I1 \\ 
    \citet{Santini_a_23_decomp_HGS} & CVRP & r-f, c-s & s & CVRP & - & - & HGS \\ 
    \citet{Schneider_15_tbra_tw} & VRPTW & c-f, r-s & std & TSP & - & - & TS \\ 
    \citet{Vidal_13_HGS-TW}  & VRPTW & r-f, c-s & s & VRPTW & rules & st & GA \\ 
    \citet{Wong_84_del_area} & VRPTW & c-f, r-s & d & TSP & - & - & 3-opt tour \\ 
    \citet{Yucenur_11_clst_ga_mdvrp} & MDVRP & c-f & s & CVRP & - & - & - \\ 
    \midrule
    \midrule
    Our approach & VRPTW & c-f, r-s & std & VRPTW & data & std & HGS-TW \\ 
    \bottomrule
\end{tabular}

        \begin{tablenotes}
            \item[1] Features: s - spatial, t - temporal, d - demand
            \item[2] Solution algorithms: B\&b - branch \& bound, SA+ILS - simulated annealing + iterative LS
        \end{tablenotes}
    \end{threeparttable}
\end{table}

\section{Problem statement}\label{prob_statement}
We briefly introduce the well-known 3-index formulation of the VRPTW following \cite{Toth_2014_vrp} to define relevant notations and equations for the description of the DRI. We consider a complete graph $G = (V, E)$. The set $V$ includes all vertices of $G$. The vertices are grouped into the subset $V_{c} \subseteq V$, which contains $n$ customer vertices that have to be visited, and the subset $\{w\} \subseteq V$, which consists of the depot vertex $w$.
An edge $e_{i,j} \in E$ connects a pair of vertices $i, j \in V$ and is associated by a cost function $c_{i,j} = C(e_{i,j})$ accruing when a vehicle $k$ directly travels from $i$ to $j$.
The fleet $K$, a set of $m$ vehicles, is used to distribute goods stored at the depot to all customers. The capacity $Q$, which limits the load shipped by a vehicle $k$, is the same for all vehicles in K.
The demand of a customer $i$ is defined as $d_i$ and varies over customers in $V_{c}$. 
The objective (\ref{eq:of_vrp}) minimizes the total travel costs. The binary decision variable $x_{i,j,k}$ indicates if vehicle $k$ uses an edge $e_{i,j}$. Then $x_{i,j,k} = 1$, otherwise $x_{i,j,k} = 0$.
\begin{equation}
    \min \sum_{i,j\in V}\sum_{k \in K} c_{i,j} \cdot x_{i,j,k}
    \label{eq:of_vrp}
\end{equation}
The fleet size, the limited load volume of vehicles, and demand fulfillment are considered in the constraints (\ref{eq:cons_fleet_size}) - (\ref{eq:cons_veh_capa}). The binary decision variable $y_{i,k}$ states if a vertex $i$ is assigned to a vehicle $k$. Then $y_{i,k} = 1$, otherwise $y_{i,k} = 0$.
\begin{equation}
    \sum_{k \in K} y_{w,k} \leq m
    \label{eq:cons_fleet_size}
\end{equation}
\begin{equation}
    \sum_{k \in K} y_{i,k} = 1 \enspace \forall \enspace i \in V_{c}
    \label{eq:cons_node_to_route}
\end{equation}
\begin{equation}
    \sum_{i \in V_{c}} d_i \cdot y_{i,k} \leq Q \enspace \forall \enspace k \in K
    \label{eq:cons_veh_capa}
\end{equation}
All vehicles start and end their route at the depot (\ref{eq:cons_fleet_size}), and every customer vertex is assigned to exactly one vehicle (\ref{eq:cons_node_to_route}). (\ref{eq:cons_veh_capa}) ensures that the vehicle's load does not exceed its capacity.
In the VRPTW, temporal information is provided for all vertices in $V$. The service time $s_i$ is needed to fulfill the order request $d_i$ of a customer $i$, and the time interval $[e_{i}, l_{i}]$ restricts a customer's availability to a specific time window. Here, $e_i$ defines the earliest possible start time of the service, and $l_i$ is the latest allowed arrival time at a customer location. The time window of the depot $[e_w, l_w]$ defines the operational period's beginning and closing. The time variable $T_{i,k}$ defines the start of the service of $k$ at $i$. Additionally, for all edges in $E$, the corresponding travel time $t_{i,j}$ is given.
\begin{equation}
    e_j \leq T_{j,k} \leq l_j \enspace \forall \enspace k \in K, j \in V
    \label{eq:cons_cst_tw}
\end{equation}  
\begin{equation}
    x_{i,j,k}(T_{i,k} + s_i + t_{i,j}) \leq T_{j,k} \enspace \forall \enspace k \in K, i,j \in V
    \label{eq:cons_start_of_service_logic}
\end{equation}
Delivery must start within a customer's time window $[e_{j}, l_{j}]$ (\ref{eq:cons_cst_tw}). $T_{j,k}$ depends on the start of delivery and required service at the previous customer $i$ plus the travel time from $i$ to $j$. (\ref{eq:cons_start_of_service_logic}).
The solution to a VRPTW instance $\mathfrak{I}$ is the set of routes $\mathfrak{R_{I}} = \{R\}$. A route $R$ is a set that represents a closed circuit that starts and ends at $w$ and visits at least one customer vertex $i \in V_{c}$. Each route is assigned a single vehicle $k$. The set $V_{R} \subseteq V_{c}$ includes all customer vertices visited in route $R$. A route $R$ is considered feasible if it does not violate (\ref{eq:cons_veh_capa}) and (\ref{eq:cons_cst_tw}). A vehicle may wait at a delivery point $i$ if it arrives before the customer is available, hence $T_{i,k} < e_i$. However, the presented objective function (\ref{eq:of_vrp}) of the VRPTW only considers the direct costs of the total distance traveled based on the sum of $c_{i,j}$ of all routes of the solution $\mathfrak{R_{I}}$.
Let $\mathfrak{P}_{p}$ be a sub-VRPTW of $\mathfrak{I}$. Then, the overall solution is defined as the set of the individual solutions of all $q$ subproblems, i.e, $\mathfrak{R_{I}} = \{\mathfrak{R}_{p}\}$ where $p = 1,..., q$.

\section{Overview of the DRI}\label{sol_frame}
\begin{figure}[H]
    \centering
\usetikzlibrary{shapes.geometric, arrows}
\tikzstyle{vrp_lrg} = [rectangle, rounded corners, minimum height=1.cm, text centered, draw=black]
\tikzstyle{clst_action} = [rectangle, minimum width=1cm, minimum height=1.cm, text centered, draw=black, fill=gray!10]
\tikzstyle{vrp_sml} = [rectangle, rounded corners, minimum width=1.5cm, minimum height=1cm, text centered, draw=black, fill=gray!25]
\tikzstyle{...} = [rectangle, minimum width=2cm, minimum height=1cm, text centered]
\tikzstyle{prc_step_D} = [rectangle, rounded corners, dashed, minimum width=6.8cm, minimum height=3.75cm, draw=black]
\tikzstyle{prc_step_R} = [rectangle, rounded corners, dashed, minimum width=1.8cm, minimum height=3.75cm, draw=black]
\tikzstyle{prc_step_I} = [rectangle, rounded corners, dashed, minimum width=5.8cm, minimum height=3.75cm, draw=black]
\tikzstyle{slv_action} = [rectangle, minimum width=1cm, minimum height=1cm, text centered, draw=black, fill=gray!10]
\tikzstyle{ls_action} = [rectangle, minimum width=1.5cm, minimum height=1cm, text centered, draw=black, fill=gray!10]
\tikzstyle{arrow} = [thick,->,>=stealth]
    \begin{tikzpicture}[node distance=2cm, font=\fontsize{9}{10}\selectfont]
        \node (vrp_obj) [vrp_lrg, align=center] {VRPTW\\$\mathfrak{I}$};
        \node (clst_stp) [prc_step_D, label=below:$decomposition$, left of=vrp_obj, xshift = 4.4cm] {};
        \node (clst_a) [clst_action, right of=vrp_obj, xshift = 0.67cm, yshift=0cm, align=center] {customer-based\\clustering};
        \node (sub_vrp_1) [vrp_sml, right of=clst_a, xshift = 0.2cm, yshift=1.25cm] {$\mathfrak{P}_1$};
        \node (solve_1) [slv_action, right of=sub_vrp_1, xshift = 0.2cm] {solving};
        \node (route_pln_1) [vrp_sml, right of=solve_1, xshift = 0.2cm, align=center] {$\mathfrak{R}_1$};
        \node (sub_vrp_2) [vrp_sml, right of=clst_a, xshift = 0.2cm, yshift=0cm] {...};
        \node (solve_2) [slv_action, right of=sub_vrp_2, xshift = 0.2cm] {solving};
        \node (route_pln_2) [vrp_sml, right of=solve_2, xshift = 0.2cm] {...};
        \node (sub_vrp_3) [vrp_sml, right of=clst_a, xshift = 0.2cm,  yshift=-1.25cm] {$\mathfrak{P}_q$};
        \node (solve_3) [slv_action, right of=sub_vrp_3, xshift = 0.2cm] {solving};
        \node (route_pln_3) [vrp_sml, right of=solve_3, xshift = 0.2cm, align=center] {$\mathfrak{R}_q$};
        \node (clst_stp) [prc_step_R, label=below:$routing$, right of=sub_vrp_2, xshift= 0.17cm] {};
        \node (ls_a) [ls_action, right of=route_pln_2, xshift = 0.25cm, yshift=0cm, align=center] {data-based\\local search};
        \node (f_route_pln) [vrp_lrg, align=center, right of=ls_a, xshift = 0.2cm] {$\mathfrak{R_{I}}$};
        \node (clst_stp) [prc_step_I, label=below:$improvement$, right of=solve_2, xshift= 2.23cm] {};
        \draw [arrow] (vrp_obj) -- (clst_a);
        \draw [arrow] (clst_a) -- (sub_vrp_1);
        \draw [arrow] (clst_a) -- (sub_vrp_2);
        \draw [arrow] (clst_a) -- (sub_vrp_3);
        \draw [arrow] (sub_vrp_1) -- (solve_1);
        \draw [arrow] (sub_vrp_2) -- (solve_2);
        \draw [arrow] (sub_vrp_3) -- (solve_3);
        \draw [arrow] (solve_1) -- (route_pln_1);
        \draw [arrow] (solve_2) -- (route_pln_2);
        \draw [arrow] (solve_3) -- (route_pln_3);
        \draw [arrow] (route_pln_1) -- (ls_a);
        \draw [arrow] (route_pln_2) -- (ls_a);
        \draw [arrow] (route_pln_3) -- (ls_a);
        \draw [arrow] (ls_a) -- (f_route_pln);
    \end{tikzpicture}
    \caption{Overview of algorithmic framework: Decompose first, route second, improve third}
    \label{fig:high_lvl_alg_flow}
\end{figure}
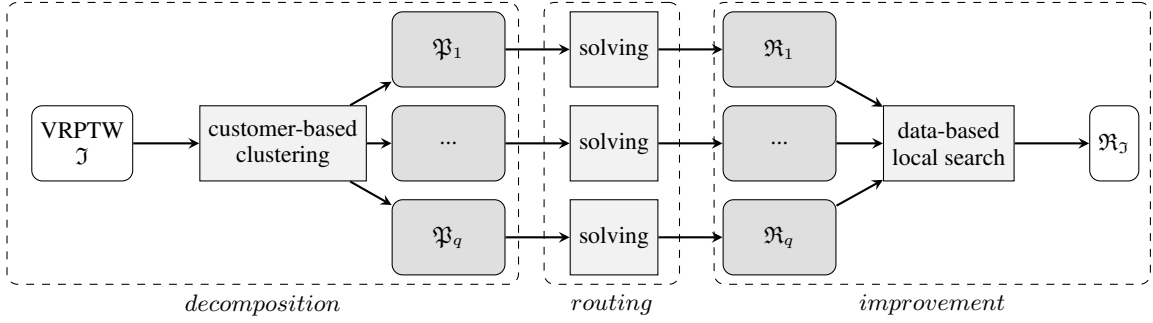

Our approach is a three-step procedure that decomposes and solves a routing problem, and improves its solution. In the \textit{decomposition phase}, we split $\mathfrak{I}$ based on spatial, temporal, and demand features of its vertices and edges of the underlying graph structure. An unsupervised clustering method groups the customer vertices into subsets. Each cluster represents a sub-VRPTW $\mathfrak{P}_{p}$ of the same type as $\mathfrak{I}$ by adding fleet and depot data. The depot is duplicated for every $\mathfrak{P}_{p}$, and vehicles are initially allocated based on the customer demands of the clusters in relation to the overall demand. 
In the \textit{routing phase}, the resulting subproblems are solved independently by a suitable method.
Subsequently, in the \textit{improvement phase}, the set of the individual route plans $\{\mathfrak{R}_{1},\ldots,\mathfrak{R}_{q}\}$ are improved using LS. Here, we focus on resolving unfavorable clustering decisions at the perimeters of the subproblems.
The framework is built to solve large-scale variants of VRPTW, but its modular and generic design may enable its application to other related routing problems. The process flow of the algorithmic framework is illustrated in Figure \ref{fig:high_lvl_alg_flow}.
\subsection{Decomposition phase}\label{sol_frame:decomp}
The number of possible routing options increases exponentially with the number of customers involved. To achieve scalability, the DRI disregards edges that are not contributing to a good solution (i.e., are associated with high transportation costs) or can be omitted because of capacity and temporal constraints by allocating customers into separate sub-VRPTWs. In clustering, similar objects are grouped together, and dissimilar objects are allocated to different clusters. Objects are represented by features, and a clustering metric measures their similarity. In the following, we introduce these concepts in the context of the VRPTW.
\subsubsection{Similarity metric:}\label{sim_metric}
In the DRI, the objects to be clustered are customers. A vertex $i \in V_{c}$ is represented by its spatial, temporal, and demand data (\ref{eq:feat_vec}). 
\begin{equation}
    \tau_i = (x_i,y_i,\theta_i, e_i, l_i, s_i, d_i)
    \label{eq:feat_vec}
\end{equation}
Spatial feature data are the geographical coordinates in $\mathbb{R}^2$ ($x_{i},y_{i})$ and the polar coordinate angle $\theta_i$ with respect to $w$.
This allows us to include information on the relative position of customers to the depot in clustering, even when the depot is excluded from the decomposition procedure. In accordance with \citet{Gillett_74_sweep}, we calculate $\theta_i$ for every customer vertex using (\ref{eq:node_depot_angle}) in its range (\ref{eq:theta_def_space}). 
\begin{equation}
    \theta_i = \arctan(\frac{y_i - y_w}{x_i - x_w})
    \label{eq:node_depot_angle}
\end{equation}
\begin{equation}
    \theta_i = \begin{cases}
            (-\pi, 0) & \text{if $y_i - y_w < 0$} \\
            [0,\pi] & \text{if $y_i - y_w \geq 0$}
    \end{cases}
    \label{eq:theta_def_space}
\end{equation}
Temporal features are $[e_{i}, l_{i}]$ and $s_i$. The demand feature is $d_i$.
Clustering metrics measure similarity of objects to be clustered. In the context of routing problems, a customer pair $i,j$ should be considered similar if $c_{i,j}$ is small and dissimilar if transportation costs are high. Let $\mathfrak{S}_{i,j}^{s}$ (\ref{eq:sim_spat_feat}) be the pairwise similarity of the spatial features of customers $i$ and $j$. Then, assuming that transportation costs positively correlate with the spatial distance between $i$ and $j$, $F(\cdot)$ approximates the underlying cost function $C(e_{i,j})$.
\begin{equation}
    \mathfrak{S}_{i,j}^{s} = F(
        x_i, y_i, \theta_i, x_j, y_j, \theta_j 
    )
    \label{eq:sim_spat_feat}
\end{equation}
If temporal and demand information are present, the exclusive consideration of spatial data to measure pairwise similarity may be delusive as the constraints of a VRP drive feasibility and quality of a solution. For the VRPTW, temporal and capacity restrictions limit flexibility in routing. In particular, the likelihood of employing an edge $e_{i,j}$ in a route $R$ is influenced by its flexibility in terms of scheduling within $R$.
The longer period $f_{i,j}$, as defined in (\ref{eq:max_buff}), between the closing time at $j$ and the latest possible departure from vertex $i$ - i.e., the sum of the opening and service time at $i$ and travel time needed if edge $e_{i,j}$ is used - the more likely $e_{i,j}$ fits into the sequence of $R$. A negative value of $f_{i,j}$ indicates that $e_{i,j}$ is infeasible. It will not appear in a feasible solution as the vehicle can only serve both customers by violating one of their time windows. Figure \ref{fig:max_flex} illustrates the maximum scheduling flexibility $f_{i,j}$.
\begin{equation}
    \label{eq:max_buff}
    f_{i,j} = l_j -(e_i + s_i + t_{i,j})
\end{equation}
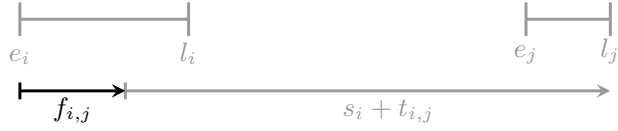
\begin{figure}
    \centering
    \begin{adjustbox}{width=0.55\linewidth}
\tikzstyle{arrow} = [line width=1pt,->,>=stealth, draw=gray!80]
\tikzstyle{arrow_highlight} = [line width=1pt,->,>=stealth, draw=black]
\tikzstyle{line_highlight} = [line width=1pt, draw=black]
\tikzstyle{line} = [line width=1pt, draw=gray!80]
\tikzstyle{node_text} = [anchor=south, text=gray!80]
\tikzstyle{node_text_highlight} = [anchor=south, text=black]

    \begin{tikzpicture}[node distance=2cm, font=\fontsize{9}{10}\selectfont]
        \draw[line] (0,0.4) -- (0,0);
        \node[node_text] at (0,-0.45) {$e_{i}$};
        \draw[line] (2,0.4) -- (2,0);
        \node[node_text] at (2,-0.45) {$l_{i}$};
        \draw[line] (0,0.2) -- (2,0.2);
        
        \draw[line] (6,0.4) -- (6,0);
        \node[node_text] at (6,-0.45) {$e_{j}$};
        \draw[line] (7,0.4) -- (7,0);
        \node[node_text] at (7,-0.45) {$l_{j}$};
        \draw[line] (6,0.2) -- (7,0.2);
        
        \draw [arrow] (1.25,-0.65) -- (7,-0.65);
        \node[node_text] at (4.37,-1.15) {$s_{i}+ t_{i,j}$};
        \draw[line] (1.25,-0.55) -- (1.25,-0.75);
        \draw [arrow_highlight] (0,-0.65) -- (1.25,-0.65);
        \node[node_text_highlight] at (0.625,-1.15) {$f_{i,j}$};
        \draw[line_highlight] (0,-0.55) -- (0,-0.75);
    \end{tikzpicture}
    \end{adjustbox}
    \caption{Illustration of the maximum arrival flexibility $f_{i,j}$ in $[e_{i},l_{i}]$}
    \label{fig:max_flex}
\end{figure}
\newpage
With increasing $f_{i,j}$, the possibility of waiting at a customer location before starting service increases. In (\ref{eq:min_h}), the minimum waiting time $h_{i,j}$ that occurs for $e_{i,j}$ is defined as the time difference between the earliest possible start date of service at $j$ and the sum of the latest possible arrival time and the time required to fulfill service at $i$ plus the time needed to travel from $i$ to $j$. The smaller $h_{i,j}$, the less time a vehicle spends waiting at $j$. Figure \ref{fig:min_wt} illustrates the minimum waiting time $h_{i,j}$.
\begin{equation}
    \label{eq:min_h}
    h_{i,j} = \max\{e_j -(l_i + s_i + t_{i,j}),0\}
\end{equation}
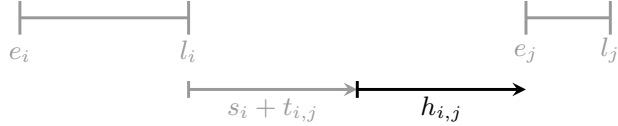
\begin{figure}[H] 
    \centering
    \begin{adjustbox}{width=0.55\linewidth}
\tikzstyle{arrow} = [line width=1pt,->,>=stealth, draw=gray!80]
\tikzstyle{arrow_highlight} = [line width=1pt,->,>=stealth, draw=black]
\tikzstyle{line_highlight} = [line width=1pt, draw=black]
\tikzstyle{line} = [line width=1pt, draw=gray!80]
\tikzstyle{node_text} = [anchor=south, text=gray!80]
\tikzstyle{node_text_highlight} = [anchor=south, text=black]

    \begin{tikzpicture}[node distance=2cm, font=\fontsize{9}{10}\selectfont]
        \draw[line] (0,0.4) -- (0,0);
        \node[node_text] at (0,-0.45) {$e_{i}$};
        \draw[line] (2,0.4) -- (2,0);
        \node[node_text] at (2,-0.45) {$l_{i}$};
        \draw[line] (0,0.2) -- (2,0.2);
        
        \draw[line] (6,0.4) -- (6,0);
        \node[node_text] at (6,-0.45) {$e_{j}$};
        \draw[line] (7,0.4) -- (7,0);
        \node[node_text] at (7,-0.45) {$l_{j}$};
        \draw[line] (6,0.2) -- (7,0.2);
        
        \draw [arrow] (2,-0.65) -- (4,-0.65);
        \node[node_text] at (3,-1.15) {$s_{i} + t_{i,j}$};
        \draw[line] (2,-0.55) -- (2,-0.75);
        \draw [arrow_highlight] (4,-0.65) -- (6,-0.65);
        \node[node_text_highlight] at (5,-1.15) {$h_{i,j}$};
        \draw[line_highlight] (4,-0.55) -- (4,-0.75);
    \end{tikzpicture}
    \end{adjustbox}
    \caption{Minimum required waiting time $h_{i,j}$ when traveling from $i$ to $j$}
    \label{fig:min_wt}
\end{figure}
The decision of which customers to be routed in sequence also depends on the capacity consumption as available choices for sequencing a customer-pair into a route decrease when the loading volume utilized by its demands increases.
The spatial-temporal-demand (STD) distance $\mathfrak{S}_{i,j}^{std}$ defined in (\ref{eq:D_std_ij}) is the pairwise spatial distance $\mathfrak{S}_{i,j}^s$ that is proportionally penalized based on depletion of temporal and capacity resources. The lower the maximum scheduling flexibility or the higher the minimum waiting time, and the more loading capacity is utilized, the higher the penalization of the spatial distance of a customer-pair. If the relative capacity consumption of customer demands is insignificant (i.e., $\frac{d_i + d_j}{Q} \simeq 0$), both time windows $[e_{i}, l_{i}]$ and $[e_{j}, l_{j}]$ are not restrictive, and the travel time $t_{i,j}$ between the two vertices is short (i.e., $h_{i,j} = 0$ and {$f_{i,j} \simeq l_w - e_w$}), then $\mathfrak{S}_{i,j}^{std} \approx \mathfrak{S}_{i,j}^s \approx C(e_{i,j})$.
\begin{equation}
    \label{eq:D_std_ij}
    \mathfrak{S}_{i,j}^{std} = \mathfrak{S}_{i,j}^{s} \cdot (2 - \frac{f_{i,j} - h_{i,j}}{l_{w} - e_{w}} + \frac{d_i + d_j}{Q})
\end{equation}
\subsubsection{Clustering algorithms:}\label{cluster_alg}
Based on $\tau_{i}$ where $i \in V_{c}$ and $\mathfrak{S}_{i,j}^{std}$, $V_{c}$ is split into $q$ different subsets $V_{c} = \{V_{p}\}$ where $p=1,\ldots,q$. The number of clusters $q$ is set so that the number of customers is within the range the chosen solution algorithm performs most efficiently in each subset. We define $q$ as a hyperparameter, as no evaluation metric exists that can be used to optimize $q$ in the context of vehicle routing. As our similarity measure is directional, i.e., $\mathfrak{S}_{i,j}^{std} \neq \mathfrak{S}_{j,i}^{std}$, but in clustering a symmetric similarity score is required, we take the minimum (\ref{eq:min_std_clst}) as clustering metric. By taking the minimum, we ensure that the similarity measure always represents the arc of a pair of vertices contributing best to the underlying routing problem.
\begin{equation}
    \overline{\mathfrak{S}}_{i,j}^{std} = \min\{\mathfrak{S}_{i,j}^{std}, \mathfrak{S}_{j,i}^{std}\}
    \label{eq:min_std_clst}
\end{equation}

For illustration of the DRI, we test three different clustering algorithms: partitional $k$-medoids and fuzzy $c$-medoids, and hierarchical agglomerative clustering. The well-known $k$-medoids clustering is used as a baseline. The fuzzy $c$-medoids approach is chosen, as it generates more relational data between subproblems from which the \textit{improvement phase} can leverage. We use the medoids versions  (i.e., the clusters are represented by an actual customer) as they allow for better interpretability of similarities between vertices based on $\overline{\mathfrak{S}}_{i,j}^{std}$.
We implement agglomerative clustering as it mimics the NN heuristic. The algorithms and their suitability for the \textit{decomposition phase} of the DRI are discussed in the following.
\paragraph{} \textbf{K-medoids} \citep{Park_a_09_kmedoids} minimizes the sum of distances from the customers of each cluster to the cluster medoid $m_p$ - i.e., the customer vertex most centrally located in the cluster. 
\begin{equation}
    \min \sum_{p=1}^q\sum_{i \in V_{p}} \overline{\mathfrak{S}}_{i,m_p}^{std}
    \label{eq:of_k_mdedoids}
\end{equation}
Initially, $q$ cluster medoids are defined that are most similar to all other vertices (\ref{eq:k_medoids_seed_calc}).
\begin{equation}
    v_i = \sum_{j \in V_{c}}\frac{\overline{\mathfrak{S}}_{i,j}^{std}}{\sum_{l \in V_{c}}\overline{\mathfrak{S}}_{j,l}^{std}} \enspace \forall \enspace i \in V_{c}  
    \label{eq:k_medoids_seed_calc}
\end{equation}
Subsequently, customers are assigned to clusters. A vertex $i$ forms the cluster $V_{p}$ with all vertices most similar to $m_p$ based on $\overline{\mathfrak{S}}_{i,m_{p}}^{std}$. Then, the cluster medoids are updated following (\ref{eq:k_medoids_m_p_update}). This procedure is repeated until no medoid changes between two consecutive iterations.
\begin{equation}
    \label{eq:k_medoids_m_p_update}
    m_p = \argmin_{i \in V_{p}}(\sum_{j \in V_{p}}\overline{\mathfrak{S}}_{i,j}^{std}),\enspace p=1,\ldots,q
\end{equation}
The pseudocode of the $k$-medoids clustering is presented in Algorithm \ref{alg:kmedoids} in the Appendix. It is a straightforward and efficient partitional clustering approach. However, as the proposed algorithm is a local heuristic, allocating customers into subsets may vary if an alternative strategy for the initial medoids selection is chosen. We address this limitation in the \textit{improvement phase} by refining customer allocations to routes in the overall solution.
\paragraph{} \textbf{Fuzzy c-medoids} is based on fuzzy c-means of \citet{Bezdek_81_fcm} and calculates the degree of membership $\mu_{i,V_{p}}$ between all a vertex-cluster combination (\ref{eq:fcm_degree_of_membership}).
\begin{equation}
    \label{eq:fcm_degree_of_membership}
    \mu_{i,V_{p}} = \frac{1}{
        \sum_{g=1}^q(\frac{
            \overline{\mathfrak{S}}_{i,m_p}^{std}
        }{
            \overline{\mathfrak{S}}_{i,m_g}^{std}
        }
        )^{\frac{2}{\kappa-1}}
    } 
\end{equation}
The matrix ${U = {(\mu_{i,V_{p}})}}$ where $i = 1,\ldots,n$ and $p = 1,\ldots,q$ holds all $\mu_{i,V_{p}}$. The objective in fuzzy c-medoids clustering is minimizing the variance of vertex features  within clusters (\ref{eq:of_fcm}). Like in \textit{$k$-medoids}, clusters are represented by their medoids. However, in fuzzy c-medoids, the vertices are not assigned to a specific cluster. The membership degree merely indicates a customer vertex's relative similarity to the cluster medoids. The parameter $\kappa$ controls this fuzziness of clusters. For $\kappa \simeq 1$, each vertex would be assigned exactly to one cluster.
\begin{equation}
    \label{eq:of_fcm}
    \min\enspace \sum_{p=1}^q \sum_{i \in V_{c}} \mu_{i,V_{p}}^{\kappa} \cdot \overline{\mathfrak{S}}_{i,m_{p}}^{std}
\end{equation}
s.t.
\begin{equation} 
    \label{eq:cons_prob_to_one_per_c}
    \sum_{p=1}^{q} \mu_{i,V_{p}} = 1 \enspace \forall \enspace i \in V_{c}
\end{equation}
\begin{equation} 
    \label{eq:cons_max_one_per_c}
    1 \leq \sum_{i=1}^{n} \mu_{i,V_{p}} \leq n-1, \enspace p=1,\ldots,q
\end{equation}
\begin{equation}
    \label{eq:cons_interval_mu}
    \mu_{i,V_{p}} \in [0, 1],  \enspace i \in V_{c},  \enspace p=1,\ldots,q
\end{equation}
(\ref{eq:cons_prob_to_one_per_c}) ensures that the sum of all degrees of memberships of a vertex $i$ is equal to 1. (\ref{eq:cons_max_one_per_c}) prevents the cluster $V_{p}$ from being empty or containing all vertices of $V_{c}$. The range of the degree of membership is given in (\ref{eq:cons_interval_mu}).
First, $U^0$ is initialized with random values. Then, each cluster's feature vector $\tau_p$ is calculated based on the weighted customer feature vectors. The $q$ cluster medoids are selected from all customers based on the STD similarity metric to the clusters' feature vector. Subsequently, $\mu_{i,V_{p}}$ is updated based on $\overline{\mathfrak{S}}_{i,m_{p}}^{std}$. The algorithm terminates if no changes are larger than a given $\epsilon$ in $U$ between two consecutive iterations. The pseudocode of the fuzzy c-medoids method is presented in Algorithm \ref{alg:f_c_m} in the Appendix.
Different to $k$-medoids, fuzzy c-medoids performs a \textit{soft} assignment of customers to clusters. In the \textit{improvement phase}, this is beneficial to identify potential routing improvements faster. Since $U$ determines the similarity of each customer to every cluster with regards to $\overline{\mathfrak{S}}_{i,m_p}^{std}$, in LS, moves can be more efficiently chosen based on the proximity data of customer vertices to other sub-routing problems.
In the \textit{routing phase}, customers must be allocated to distinct subsets to create a sub-VRPTW $\mathfrak{P}_p$. Therefore, we assign each vertex in $V_{c}$ to cluster $V_{p^*}$ where $p^* = \argmax_{p}(\mu_{i, V_{p}})$ and $p=1,\ldots,q$.
\paragraph{} \textbf{Agglomerative clustering} \citep{Anderberg_73_clust} is a bottom-up hierarchical clustering approach. Initially, each customer $i \in V_{c}$ represents its own cluster $V_{p}$ where $p=1,\ldots,n$, i.e., $|V_{p}|=1$. Then, larger customer subsets are created by iteratively combining clusters until all vertices of $V_{c}$ are allocated to $q$ clusters. In every iteration, the cluster-pair ($V_{p}$,$V_{g}$) is merged that is most similar with regards to the STD similarity metric (\ref{eq:of_agglom}). In case of a tie, $(V_{p},V_{g})$ is chosen randomly with equal probability.
\begin{equation}
    \label{eq:of_agglom}
    (V_{p},V_{g}) = \argmin_{p,g}(\overline{\mathfrak{S}}_{V_{p},V_{g}}^{std}), \enspace p=1,\ldots,q, \enspace g=1,\ldots,q, \enspace p \neq g
\end{equation}
Here, $\overline{\mathfrak{S}}_{V_{p},V_{g}}^{std}$ depends on the linkage method that determines the customer vertices representing the clusters. In \textit{single} linkage (\ref{eq:dist_sl}), the STD similarity of $V_{p}$ and $V_{g}$ equals the most similar vertex-pair $i \in V_{p}$ and $j \in V_{g}$.
\begin{equation}
    \label{eq:dist_sl}
   \overline{\mathfrak{S}}_{V_{p},V_{g}}^{std} = \min_{i\in V_{p},j \in  V_{g}}(\overline{\mathfrak{S}}_{i,j}^{std})
\end{equation}
Conversely, in \textit{complete} linkage (\ref{eq:dist_cl}), $\overline{\mathfrak{S}}_{V_{p},V_{g}}^{std}$ is equal to the similarity value of the most dissimilar vertex-pair $i \in V_{p}$ and $j \in V_{g}$ between the $V_{p}$ and $V_{g}$. 
\begin{equation}
    \label{eq:dist_cl}
    \overline{\mathfrak{S}}_{V_{p},V_{g}}^{std} = \max_{i\in V_{p},j \in  V_{g}}(\overline{\mathfrak{S}}_{i,j}^{std})
\end{equation}
In \textit{average} linkage (\ref{eq:dist_al}), the average of $\overline{\mathfrak{S}}_{i,j}^{std}$ between all vertex-pair combinations $i \in V_{p}$ and $j \in V_{g}$ is defined as $\overline{\mathfrak{S}}_{V_{p},V_{g}}^{std}$. 
\begin{equation}
    \label{eq:dist_al}
    \overline{\mathfrak{S}}_{V_{p},V_{g}}^{std} = \frac{1}{|V_{p}|\cdot|V_{g}|} \sum_{i\in V_{p}}\sum_{j \in V_{g}}\overline{\mathfrak{S}}_{i,j}^{std}
\end{equation}
The pseudocode of agglomerative clustering is presented in Algorithm \ref{alg:ac} in the Appendix. Unlike $k$-medoid and fuzzy c-medoids clustering, agglomerative clustering does not depend on any initialization strategy. The formation of clusters solely depends on the chosen linkage method for the hierarchical approach. Single linkage shares similarities with the NN heuristic for routing. However, this greedy approach may unevenly allocate customers to clusters, and thus, sub-routing problems would vary significantly in size. We address this issue by adjusting the runtime limit for the different sub-VRPTWs based on the number of customers included.

\subsection{Routing phase}\label{sol_frame:route}
The stand-alone routing problems $\mathfrak{P}_{p}$ ($p=1,\ldots,q$) are created by duplicating the depot $w$ for every subproblem and adding a customer subset $V_{p}$ to each $\mathfrak{P}_{p}$.
The total number of available vehicles $m$ of $\mathfrak{I}$ is split across the subproblems in relation to the ratio of their demands and the total demand of the complete problem. A fleet $K_p$ assigned to $\mathfrak{P}_{p}$ is calculated using the ceiling function (\ref{eq:veh_per_clst}). This fleet allocation strategy is most natural, assuming $K$ is larger than the minimum required fleet.
\begin{equation}
    \label{eq:veh_per_clst}
    K_p = \lceil K \cdot \frac{\sum_{i \in V_{p}}d_i}{\sum_{i \in V_{c}}d_i} \rceil 
\end{equation}
In the \textit{routing phase}, each $\mathfrak{P}_{p}$ is solved separately. Ultimately, any suitable solution algorithm can be applied to generate a tour plan $\mathfrak{R}_{p}$. 
The overarching goal is to find near-optimal solutions with minimum required runtime. Thus, we use a state-of-the-art metaheuristics in terms of solution quality and convergence rate for small and mid-size instances of the VRPTW
Let $\Theta$ be the overall runtime limit of the DRI and $\nu$ be the time needed to create the set of subproblems $\{\mathfrak{P}_{1}, \ldots, \mathfrak{P}_{q}\}$. Then, the remaining time $\Delta$ in (\ref{eq:rt_ri}) is split between the \textit{routing} (\ref{eq:rt_r}) and \textit{improvement phase} (\ref{eq:rt_i}) based on the hyperparameter $\alpha$.
\begin{equation}
    \label{eq:rt_ri}
    \Delta = \Theta - \nu
\end{equation}
\begin{equation}
    \label{eq:rt_r}
    \Omega = \alpha \cdot \Delta
\end{equation}
\begin{equation}
    \label{eq:rt_i}
    \Upsilon = (1 - \alpha) \cdot \Delta
\end{equation}
When solved sequentially, we allocate $\Omega$ to the subproblems in relation to the ratio of their size, i.e., the number of customers in the subset $p$, and the size of the overall problem $n$ using the floor function (\ref{eq:rt_r_split}). Thus, the more customers in a subproblem, the more computation time is assigned in the \textit{routing phase}.
\begin{equation}
    \label{eq:rt_r_split}
    \Omega_{p} = \lfloor \Omega \cdot \frac{V_{p}}{n} \rfloor
\end{equation}
\subsection{Improvement phase}\label{sol_frame:improve}
In the final step of the DRI, the \textit{improvement phase}, we apply LS to efficiently eradicate potential inefficient routes caused by the split of the customers in the \textit{decomposition phase}.
\subsubsection{Evaluating the subproblems solution quality:} We indicate whether a customer subset was solved efficiently separate from the others and which may improve when routing decisions are reconsidered by combining the route plans of the subproblems by the following measurements.
Let $Z_{\mathfrak{R}_{p}}$ (\ref{eq:R_costs}) be the total travel costs of route set $\mathfrak{R}_{p}$ and let $\overline{Z}_{\mathfrak{R}_{p}}$ (\ref{eq:avg_cost_per_route}) be the average travel costs per route of $\mathfrak{P}_{p}$. Further, let $\mathfrak{u}_{R}$ be the utilization of a route $R \in \mathfrak{R}_{p}$ (\ref{eq:r_util}).

\begin{equation}
    \label{eq:R_costs}
    Z_{\mathfrak{R}_{p}} = \sum_{R \in \mathfrak{R}_{p}}\sum_{e_{i,j} \in R} c_{i,j}
\end{equation}

\begin{equation}
    \label{eq:avg_cost_per_route}
    \overline{Z}_{\mathfrak{R}_{p}} = \frac{1}{|\mathfrak{R}_{p}|}Z_{\mathfrak{R}_{p}}
\end{equation}
\begin{equation}
    \label{eq:r_util}
    \mathfrak{u}_{R} = \frac{\sum_{i \in R}d_i}{Q}
\end{equation}
The higher the average route costs of a solution, the more customers are involved or the longer the distances between customers. Therefore, it is less likely that the separate solution of the subproblem is efficiently contributing to the overall solution. The lower a route's utilization, the more likely it can be removed when its customers are assigned to other vehicles. Thus, LS tries to improve $R \in \mathfrak{R}_{p}$ with $\argmax_{p}(\overline{Z}_{\mathfrak{R}_{p}})$ first. The routes are ordered ascendingly based on $\mathfrak{u}_{R}$. Subsequently, all routes of $\mathfrak{R}_{p}$ with the next highest value of $\overline{Z}_{\mathfrak{R}_{p}}$ are similarly ordered. Consequently, the last route LS is applied on is $R = \argmax_{R}(\mathfrak{u}_{R})$ of $\mathfrak{R}_{p}$ with $\argmin_{p}(\overline{Z}_{\mathfrak{R}_{p}})$.

\subsubsection{Relational data of subproblems and customers:}
Assuming that the \textit{routing phase} yields individual near-optimal solutions for all $\mathfrak{P}_{p}$, LS should focus on inter-route improvements between subproblems. Inter-route operators move a customer or a sequence of customers from one route to another, i.e., multiple routes are modified simultaneously. We prune LS moves to potentially unfavorable routing decisions between similar subproblems based on relational data obtained in the \textit{decomposition phase}.
An inter-route operation is only applied on a route pair $(R,R')$ where $R \in \mathfrak{R}_{p}$ and $R' \in \mathfrak{R}_{g}$ and $p \neq g$. Further, $\mathfrak{P}_{g}$ must lay in the \textit{vicinity} $\Phi_{p}$ of $\mathfrak{P}_{p}$. Here, $\Phi_{p}$ is defined as the set of the $\phi$ nearest subproblems to $\mathfrak{P}_{p}$, i.e., ${\Phi_{p}=\{\mathfrak{P}_{p},g=1,..,\phi\}}$. The distance between two subproblems $\mathfrak{P}_{p}$ and $\mathfrak{P}_{g}$ is measured based on $\overline{\mathfrak{S}}_{V_{p},V_{g}}^{std}$. 
On the customer level, the vicinity $\Phi_{i}$ is a set of the $\varphi$ most similar vertices of $i$ based on $\overline{\mathfrak{S}}_{i,j}^{std}$.
If fuzzy c-medoids is used in the \textit{decomposition phase}, LS is also educated based on the degree of membership $\mu_{i,V_{p}}$ where $i \in V_{p}, p=1,\ldots,q$. A vertex $i \in R$ is only applicable for an inter-route move if $\mu_{i,V_{p}} \leq \rho$ where $\rho = [0,1]$. Consequently, moves are more focused on the borderlines of subproblems. 

\subsubsection{Local search:}
For inter-route operations, we exemplary use \textit{cross-over}, \textit{relocate}, \textit{swap}, and the 2opt-heuristic. These operators have proven their effectiveness in previous studies \citep[e.g.,][]{Vidal_12_HGS_org, Arnold_19_v_large_scale}.
\textit{Cross-over} cuts the route-pair at a certain position and swaps the tails of the routes. The \textit{relocate} operator removes a vertex from one route and inserts it at a specific position into another. In a \textit{swap} move, two customer vertices of different routes switch places.
If an improvement is found, routes will be updated as their vertex assignment and sequencing have changed. After every successful application of an inter-route operator, an intra-route operation is applied on each updated route $R^{*}$. Intra-route operators improve a single route by modifying the sequence of the customer vertices within a route. The following intra-route operators are applicable in the DRI: \textit{swap} and the 2opt-heuristic. The two search strategies \textit{first-descent} and \textit{steepest-descent} are available for all LS operations. The first strategy accepts the first improvement of a LS operation, returns the changes in the routing, and continues with another operation until it finds no further improvement. The latter evaluates all available moves of an operation and returns the overall best before continuing with another LS operation in the next iteration.
LS either stops if no further improvement can be found for all possible moves or its runtime limit $\Upsilon$ is reached. At the end of the \textit{improvement phase}, all routes of the former subproblems are combined to the final tour plan of the overall problem, i.e., $\mathfrak{R}_{\mathfrak{I}} = \{\mathfrak{R}_{1}^{*}, \ldots, \mathfrak{R}_{q}^{*}\}$.

\begin{figure}[ht] 
    \centering
    \begin{adjustbox}{width=0.43\linewidth}
\usetikzlibrary{shapes.geometric, arrows}

    \begin{tikzpicture}
        \node[circle, white, fill=white] (c) at (0.2,0.2) {};
        \node[circle, white, fill=white] (c) at (0.2,9.8) {};
        \node[circle, white, fill=white] (c) at (9.8,9.8) {};
        \node[circle, white, fill=white] (c) at (9.8,0.2) {};
        \node [rectangle, draw=black, thick, minimum size = 0.4cm] (depot) at (5,5) {D};
        \node[circle, black, fill=black] (c1) at (1,1) {};
        \node[circle, black, fill=black] (c2) at (1,5) {};
        \node[circle, black, fill=black] (c3) at (2,4) {};
        \node[circle, black, fill=black] (c4) at (3,2) {};
        \node[circle, black, fill=black] (c5) at (4,3) {};
        \node[circle, black, fill=black] (c6) at (3,5.2) {};
        \node[circle, black, fill=black] (c7) at (8,7) {};
        \node[circle, black, fill=black] (c8) at (6,7) {};
        \node[circle, black, fill=black] (c9) at (9,6) {};
        \node[circle, black, fill=black] (c10) at (5,8) {};
        \node[circle, black, fill=black] (c11) at (9,9) {};
        \node[circle, black, fill=black] (c12) at (5,6) {};
        \node[circle, black, fill=black] (c13) at (7.5,5) {};
        \node[circle, gray!50, fill=gray!50] (c14) at (9,1.5) {};
        \node[circle, gray!50, fill=gray!50] (c15) at (7,2.5) {};
        \node[circle, gray!50, fill=gray!50] (c17) at (7.5,1) {};
        \node[circle, gray!50, fill=gray!50] (c18) at (6,4) {};
        \node[circle, gray!50, fill=gray!50] (c16) at (8,3.5) {};
        \node[circle, gray!50, fill=gray!50] (c19) at (5.33,1.75) {};
        \node[circle, black, fill=black] (c20) at (1,8.5) {};
        \node[diamond, black, fill=black] (c21) at (3,8) {};
        \node[circle, black, fill=black] (c22) at (1.5,9) {};
        \node[circle, black, fill=black] (c23) at (2.5,8) {};
        \node[diamond, black, fill=black] (c24) at (1,7) {};
        \node[circle, black, fill=black] (c25) at (1.8,7.8) {};
        \draw [dashed, gray!75] plot[smooth cycle] coordinates {(0.7, 0.7) (0.5, 5.1) (3.5, 5.4) (4.3, 2.85)};
        \node [rectangle, draw=white, minimum size = 0.3cm] (P1) at (1,2.5) {$\mathfrak{P}_1$};
        \draw [dashed, draw=gray!50] plot[smooth cycle] coordinates {(9.2, 9.2) (9.1, 5.5) (7.5, 4.77) (4.8, 5.85) (4.8, 8.35)};
        \node [rectangle, draw=white, minimum size = 0.3cm] (P2) at (7,7.6) {$\mathfrak{P}_2$};
        \draw [dashed, draw=gray!85] plot[smooth cycle] coordinates {(9.2,1) (5.2,1.25) (5.33, 4.25) (8, 4)};
        \node [rectangle, draw=white, minimum size = 0.3cm, text=gray!50] (P3) at (8.2,1.2) {$\mathfrak{P}_3$};
        \draw [dashed, draw=black] plot[smooth cycle] coordinates {(1,9.2) (3.25,9) (3, 7.5) (0.7, 6.7)};
        \node [rectangle, draw=white, minimum size = 0.3cm] (P4) at (2.2,8.7) {$\mathfrak{P}_4$};
        \node [rectangle, draw=black, thick, minimum size = 0.4cm] (cc) at (2,4) {};
        \node [rectangle, draw=white, thick, minimum size = 0.2cm] (cc) at (1.4,4) {$m_1$};
        \node [rectangle, draw=black, thick, minimum size = 0.4cm] (cc) at (8,7) {};
        \node [rectangle, draw=white, thick, minimum size = 0.2cm] (cc) at (7.4,7) {$m_2$};
        \node [rectangle, draw=gray!50, thick, minimum size = 0.4cm] (cc) at (7,2.5) {};
        \node [rectangle, draw=white, thick, minimum size = 0.2cm, text=gray!50] (cc) at (7.6,2.5) {$m_3$};
        \node [rectangle, draw=black, thick, minimum size = 0.4cm] (cc) at (1.8,7.8) {};
        \node [rectangle, draw=white, thick, minimum size = 0.2cm] (cc) at (1.8,7.3) {$m_4$};
        \draw (depot) -- (c5); 
        \draw (c5) -- (c4);
        \draw (c4) -- (c1);
        \draw (c1) -- (depot);
        \draw (depot) -- (c3);
        \draw (c3) -- (c2);
        \draw (c2) -- (c6);
        \draw (c6) -- (depot);
        \draw (depot) -- (c12);
        \draw (c12) -- (c8);
        \draw (c8) -- (c10);
        \draw (c10) -- (c11);
        \draw (c11) -- (c7);
        \draw (c7) -- (depot);
        \draw (depot) -- (c13);
        \draw (c13) -- (c9);
        \draw (c9) -- (depot);
        \draw[opacity=0.3] (depot) -- (c16);
        \draw[opacity=0.3] (c16) -- (c14);
        \draw[opacity=0.3] (c14) -- (c15);
        \draw[opacity=0.3] (c15) -- (c18);
        \draw[opacity=0.3] (c18) -- (depot);
        \draw[opacity=0.3] (depot) -- (c17);
        \draw[opacity=0.3] (c17) -- (c19);
        \draw[opacity=0.3] (c19) -- (depot);
        \draw (depot) -- (c24);
        \draw (c24) -- (c20);
        \draw (c20) -- (c22);
        \draw (c22) -- (c25);
        \draw (c25) -- (depot);
        \draw (depot) -- (c21);
        \draw (c21) -- (c23);
        \draw (c23) -- (depot);
        \draw [dotted, draw=black] plot[smooth cycle] coordinates {(1.5,9.5) (9.2,9.4) (9.1, 4.9) (6.5, 4.8) (4.5, 5.5) (4.33, 2) (0.5, 0.8) (0.3, 7)};
        \node [rectangle, dotted, draw=black, thick, minimum size = 0.3cm] (cc) at (4,9) {$\Phi_4$};
        \node [rectangle, minimum size = 0.3cm] (r) at (3.67,7.6) {$R$};
        \node [rectangle, minimum size = 0.3cm] (i) at (3,8.5) {$i$};
        \node [rectangle, minimum size = 0.3cm] (r) at (6.5,6.5) {$R'$};
        \node [rectangle, minimum size = 0.3cm] (j) at (5,6.5) {$j$};
        \draw [dashed, draw=black] plot[smooth cycle] coordinates {(4.5,6) (4.5,8.2) (5.4, 8.2) (5.4, 6)};
        \node [rectangle, dotted, draw=black, thick, minimum size = 0.3cm] (cc) at (4.9,7.3) {$\Phi_i$};
         \node [rectangle, draw=black, thick, minimum size = 10cm] (f) at (5,5) {};
        
    \end{tikzpicture}
    \end{adjustbox}
    \begin{adjustbox}{width=0.43\linewidth}
\usetikzlibrary{shapes.geometric, arrows}

    \begin{tikzpicture}
        \node[circle, white, fill=white] (c) at (0.2,0.2) {};
        \node[circle, white, fill=white] (c) at (0.2,9.8) {};
        \node[circle, white, fill=white] (c) at (9.8,9.8) {};
        \node[circle, white, fill=white] (c) at (9.8,0.2) {};
        \node [rectangle, draw=black, thick, minimum size = 0.4cm] (depot) at (5,5) {D};
        \node[circle, black, fill=black] (c1) at (1,1) {};
        \node[circle, black, fill=black] (c2) at (1,5) {};
        \node[circle, black, fill=black] (c3) at (2,4) {};
        \node[circle, black, fill=black] (c4) at (3,2) {};
        \node[circle, black, fill=black] (c5) at (4,3) {};
        \node[circle, black, fill=black] (c6) at (3,5.2) {};
        \node[circle, black, fill=black] (c7) at (8,7) {};
        \node[circle, black, fill=black] (c8) at (6,7) {};
        \node[circle, black, fill=black] (c9) at (9,6) {};
        \node[circle, black, fill=black] (c10) at (5,8) {};
        \node[circle, black, fill=black] (c11) at (9,9) {};
        \node[circle, black, fill=black] (c12) at (5,6) {};
        \node[circle, black, fill=black] (c13) at (7.5,5) {};
        \node[circle, gray!50, fill=gray!50] (c14) at (9,1.5) {};
        \node[circle, gray!50, fill=gray!50] (c15) at (7,2.5) {};
        \node[circle, gray!50, fill=gray!50] (c17) at (7.5,1) {};
        \node[circle, gray!50, fill=gray!50] (c18) at (6,4) {};
        \node[circle, gray!50, fill=gray!50] (c16) at (8,3.5) {};
        \node[circle, gray!50, fill=gray!50] (c19) at (5.33,1.75) {};
        \node[circle, black, fill=black] (c20) at (1,8.5) {};
        \node[diamond, black, fill=black] (c21) at (3,8) {};
        \node[circle, black, fill=black] (c22) at (1.5,9) {};
        \node[circle, black, fill=black] (c23) at (2.5,8) {};
        \node[diamond, black, fill=black] (c24) at (1,7) {};
        \node[circle, black, fill=black] (c25) at (1.8,7.8) {};
        \draw [dashed, gray!75] plot[smooth cycle] coordinates {(0.7, 0.7) (0.5, 5.1) (3.5, 5.4) (4.3, 2.85)};
        \node [rectangle, draw=white, minimum size = 0.3cm] (P1) at (1,2.5) {$\mathfrak{P}_1$};
        \draw [dashed, draw=gray!50] plot[smooth cycle] coordinates {(9.2, 9.2) (9.1, 5.5) (7.5, 4.77) (4.8, 5.85) (4.8, 8.35)};
        \node [rectangle, draw=white, minimum size = 0.3cm] (P2) at (7,7.6) {$\mathfrak{P}_2$};
        \draw [dashed, draw=gray!85] plot[smooth cycle] coordinates {(9.2,1) (5.2,1.25) (5.33, 4.25) (8, 4)};
        \node [rectangle, draw=white, minimum size = 0.3cm, text=gray!50] (P3) at (8.2,1.2) {$\mathfrak{P}_3$};
        \draw [dashed, draw=black] plot[smooth cycle] coordinates {(1,9.2) (3.25,9) (3, 7.5) (0.7, 6.7)};
        \node [rectangle, draw=white, minimum size = 0.3cm] (P4) at (2.2,8.7) {$\mathfrak{P}_4$};
        \node [rectangle, draw=black, thick, minimum size = 0.4cm] (cc) at (2,4) {};
        \node [rectangle, draw=white, thick, minimum size = 0.2cm] (cc) at (1.4,4) {$m_1$};
        \node [rectangle, draw=black, thick, minimum size = 0.4cm] (cc) at (8,7) {};
        \node [rectangle, draw=white, thick, minimum size = 0.2cm] (cc) at (7.4,7) {$m_2$};
        \node [rectangle, draw=gray!50, thick, minimum size = 0.4cm] (cc) at (7,2.5) {};
        \node [rectangle, draw=white, thick, minimum size = 0.2cm, text=gray!50] (cc) at (7.6,2.5) {$m_3$};
        \node [rectangle, draw=black, thick, minimum size = 0.4cm] (cc) at (1.8,7.8) {};
        \node [rectangle, draw=white, thick, minimum size = 0.2cm] (cc) at (1.8,7.3) {$m_4$};
        \draw (depot) -- (c5); 
        \draw (c5) -- (c4);
        \draw (c4) -- (c1);
        \draw (c1) -- (depot);
        \draw (depot) -- (c3);
        \draw (c3) -- (c2);
        \draw (c2) -- (c6);
        \draw (c6) -- (depot);
        \draw (depot) -- (c8);
        \draw (c8) -- (c10);
        \draw (c10) -- (c11);
        \draw (c11) -- (c7);
        \draw (c7) -- (depot);
        \draw (depot) -- (c13);
        \draw (c13) -- (c9);
        \draw (c9) -- (depot);
        \draw[opacity=0.3] (depot) -- (c16);
        \draw[opacity=0.3] (c16) -- (c14);
        \draw[opacity=0.3] (c14) -- (c15);
        \draw[opacity=0.3] (c15) -- (c18);
        \draw[opacity=0.3] (c18) -- (depot);
        \draw[opacity=0.3] (depot) -- (c17);
        \draw[opacity=0.3] (c17) -- (c19);
        \draw[opacity=0.3] (c19) -- (depot);
        \draw (depot) -- (c24);
        \draw (c24) -- (c20);
        \draw (c20) -- (c22);
        \draw (c22) -- (c25);
        \draw (c25) -- (depot);
        
        \draw (depot) -- (c12);
        \draw (c12) -- (c21);
        \draw (c21) -- (c23);
        \draw (c23) -- (depot);
        \draw [dotted, draw=black] plot[smooth cycle] coordinates {(1.5,9.5) (9.2,9.4) (9.1, 4.9) (6.5, 4.8) (4.5, 5.5) (4.33, 2) (0.5, 0.8) (0.3, 7)};
        \node [rectangle, dotted, draw=black, thick, minimum size = 0.3cm] (cc) at (4,9) {$\Phi_4$};
        \node [rectangle, minimum size = 0.3cm] (r) at (3.67,7.7) {$R$};
        \node [rectangle, minimum size = 0.3cm] (i) at (3,8.5) {$i$};
        \node [rectangle, minimum size = 0.3cm] (r) at (6.5,6.5) {$R'$};
        \node [rectangle, minimum size = 0.3cm] (j) at (5,6.5) {$j$};
        \draw [dashed, draw=black] plot[smooth cycle] coordinates {(4.5,6) (4.5,8.2) (5.4, 8.2) (5.4, 6)};
        \node [rectangle, dotted, draw=black, thick, minimum size = 0.3cm] (cc) at (4.9,7.3) {$\Phi_i$};
         \node [rectangle, draw=black, thick, minimum size = 10cm] (f) at (5,5) {};
        
    \end{tikzpicture}
    \end{adjustbox}
    \caption{Illustration of a data-based inter-route move of $i \in R$ and $j \in R'$ within $\Phi_{4}$ and $\Phi_{i}$}
    \label{fig:vrp_ls_CO}
\end{figure}
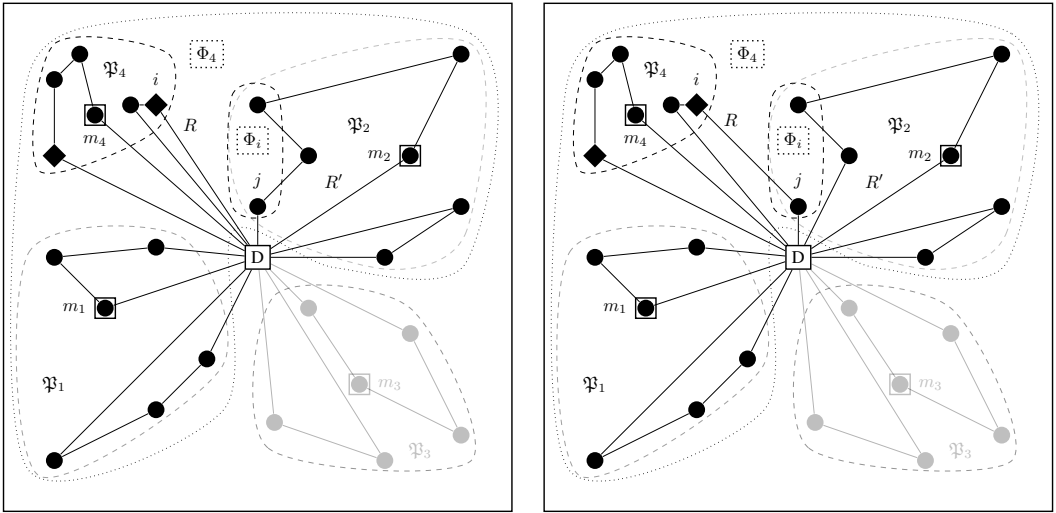

Figure \ref{fig:vrp_ls_CO} illustrates an inter-route operation for a fuzzy vertex $i$ in $R$ of $\mathfrak{P}_{4}$. Its subproblem's vicinity $\Phi_{4}$ includes the sub-VRPTWs $\mathfrak{P}_{1}$ and $\mathfrak{P}_{2}$, i.e., $\phi = 2$. Node $j$ of route $R'$ lays within both vicinities $\Phi_{4}$ and $\Phi_{i}$ where $\varphi = 2$. Thus an inter-route move between $R \in \mathfrak{P}_{4}$ and $R' \in \mathfrak{P}_{2}$ is performed to evaluate the routing if a vertex $j$ would have been assigned to $\mathfrak{P}_{4}$ rather than $\mathfrak{P}_{2}$.
\section{Computational study}\label{cs}
We perform a comprehensive numerical study to test the performance of the DRI on large-scale VRPTWs. First, we examine the impact of different problem characteristics, such as instance size, customer location distribution, and vehicle capacity, on the DRI's hyperparameters. In the \textit{decomposition phase}, the external parameter $q$ is independent of the clustering approach. \textit{Internal} parameters are $\kappa$ in fuzzy c-medoids and linkage in agglomerative clustering. In the \textit{improvement phase}, we evaluate the effect of vicinity sizes and search strategies on the efficiency of LS. Thus, we provide insights into how well the DRI can be adjusted to generate high-quality solutions for various distribution scenarios.
Further, we evaluate the impact of customer features on the solution and how the \textit{decomposition} and \textit{improvement phase} leverage from this data. In particular, we compare the introduced similarity metric $\mathfrak{S}_{i,j}^{std}$ against a baseline clustering metric and measure the effect of data-based LS against rule-based approaches.
Additionally, we investigate the influence of budgeting the computation time between the \textit{routing} and \textit{improvement phase}.
Finally, we benchmark our results against the HGS-TW implementation by \citet{Kool_22_hgs_tw_impl} (https://pyvrp.org/).
\subsection{Dataset and test environment}
We use the $180$ large-scale VRPTW instances with $600-1000$ customers of \citet{gehring_99_bm_vrptw}. The instances are distinguished concerning spatial distribution of customers in clustered (C), random (R), and random-clustered instances (RC). Further, C$1$, R$1$, and RC$1$ instances have a short operational period (i.e., a smaller time difference between the closing and opening time of the depot) and tight capacity restrictions for the vehicles. In contrast, C$2$, R$2$, and RC$2$ instances are characterized by a high vehicle capacity and an extended operational period. Thus, more customers are visited within a route, and the routes are long. The time-window scenarios range from very restrictive time windows for all customers to 75\% customer time windows with almost no limiting effect.
All experiments are run on a single thread of an Intel(R) Xeon(R) Platinum 8160 CPU 2.1 GHz processor with \textbf{2.8}GB of RAM running Ubuntu 20.04 LTS. The DRI framework is implemented in Python 3.9 and calls the HGS-TW implemented in C++ and LS operations implemented in Julia 1.8.2.
\subsection{Hyperparameters - decomposition phase}
For the autonomous analyses of the \textit{decomposition phase's} hyperparameters, we terminate the DRI after the \textit{routing phase}. The subproblems are solved using the HGS-TW. Its termination criterion is set to $5,000$ successive iterations performed without improvement to achieve sufficient convergence of the algorithm.
\subsubsection{External parameters:}
\begin{figure}[ht] 
    \centering
    \begin{adjustbox}{width=0.75\linewidth}
        \includegraphics[width=\linewidth]{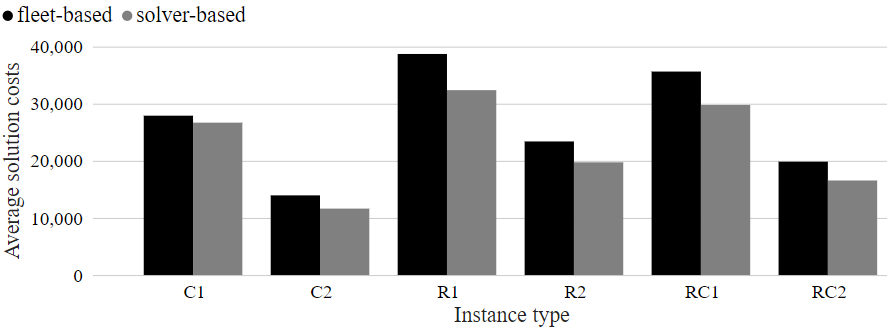}
    \end{adjustbox}
    \caption{Impact of the strategy to determine $q$ on the routing costs}
    \label{fig:no_of_clusters}
\end{figure}

The number of subproblems $q$ the VRPTW is split into directly impacts routing flexibility. Most \textit{cluster-first, route-second} approaches create subproblems that are served by a single vehicle. Thus, $q$ is determined based on the minimum required fleet size to serve all customers, i.e., fleet-based. The solver-based approach sets $q$ such that the number of customers in the subproblems can be solved efficiently, i.e., close to optimality in reasonable time, by the chosen solution algorithm. Previous studies show that the HGS-TW performs best on problems with less than $500$ customers. Figure \ref{fig:no_of_clusters} shows that the solver-based approach leads, on average, to lower costs for all instance types. The fleet-based approach fails to identify good subproblems when customers are randomly located. If customers are located in clusters, and routes are short, i.e., C1 instances, more subproblems lead only to slightly worse solutions.

\begin{figure}[ht]
    \centering
    \begin{adjustbox}{width=0.75\linewidth}
        \includegraphics[width=\linewidth]{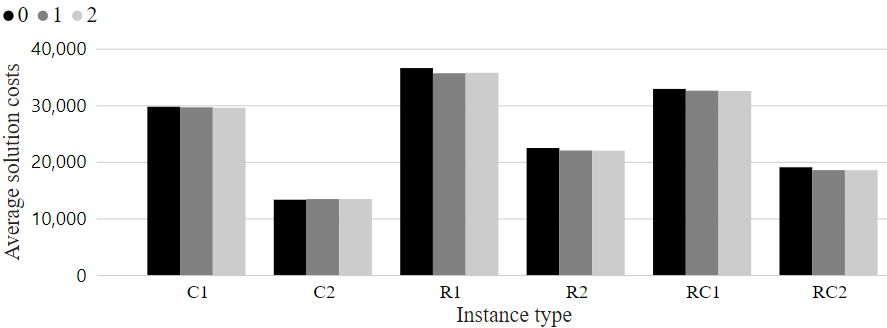}
    \end{adjustbox}
    \caption{Impact of the vertex depot angle weighting factor $\lambda = \{0,1,2\}$ on the routing costs}
    \label{fig:hybrid_weight_spatial}
\end{figure}

The pair-wise spatial similarity $\mathfrak{S}_{i,j}^{s}$ estimates the travel cost function $C(e_{i,j})$ based on the customers' spatial features. In the \citet{gehring_99_bm_vrptw} benchmark instances, $C(e_{i,j})$ is equal to the Euclidean distance between a vertex pair, and $c_{i,j} = t_{i,j}$. Accordingly, we specify $\mathfrak{S}_{i,j}^{s}$ as follows.
\begin{equation}
    \mathfrak{S}_{i,j}^{s} = \sqrt{(x_j - x_i)^2 + (y_j - y_i)^2 + \lambda \cdot(\theta_j - \theta_i)^2}
    \label{eq:dist_spat_eucl_3dim}
\end{equation}
The parameter $\lambda$ controls the weight of the relative position of a customer to the depot when determining spatial similarity of customers. For $\lambda = 0$, $\mathfrak{S}_{i,j}^{s} = C(e_{i,j})$. If $\lambda = 1$, the difference between the coordinates and the difference in the customer-depot angle are equally weighted. Accordingly, if $\lambda = 2$, the difference in customers' polar angles to the depot is given twice the importance compared to the distance between their coordinates. Figure \ref{fig:hybrid_weight_spatial} shows that adding relative depot information, on average, improves the cluster creation concerning routing costs, even when spatial similarity could exactly represent travel costs. The rationale behind this is that when including the customer-depot angle, the form of the clusters is wedge-shaped, which follows the natural shape of cost-minimal routes. The relatively small improvement can be ascribed to the characteristic of the \citet{gehring_99_bm_vrptw} instances that the depot is always located in the customers' geographical center. When the depot is decentral, $\theta_{i}$ is of higher importance.
\subsubsection{Internal parameters:}
\begin{figure}[ht!] 
    \centering
    \begin{adjustbox}{width=0.88\linewidth}
        \includegraphics[width=\linewidth]{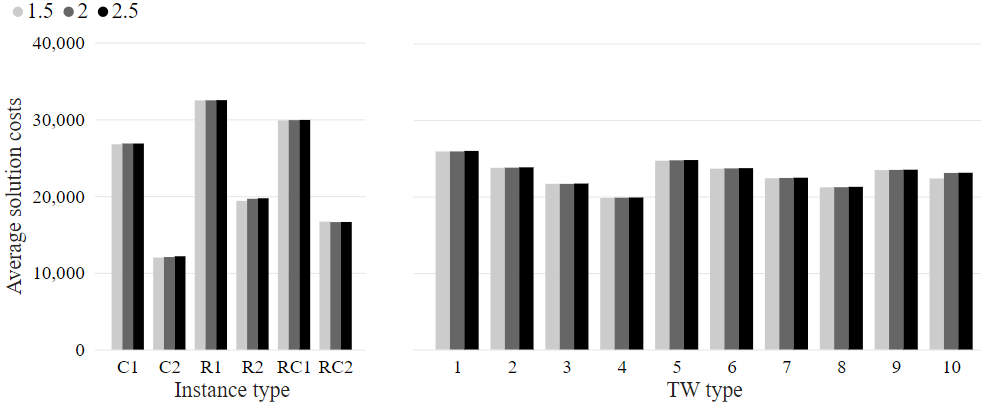}
    \end{adjustbox}
    \caption{Impact of $\kappa = \{1.5, 2, 2.5\}$ in fuzzy c-medoids on the routing costs}
    \label{fig:fuzzy_m_param}
\end{figure}

\begin{figure}[ht] %
    \centering
    \begin{adjustbox}{width=0.85\linewidth}
        \includegraphics[width=\linewidth]{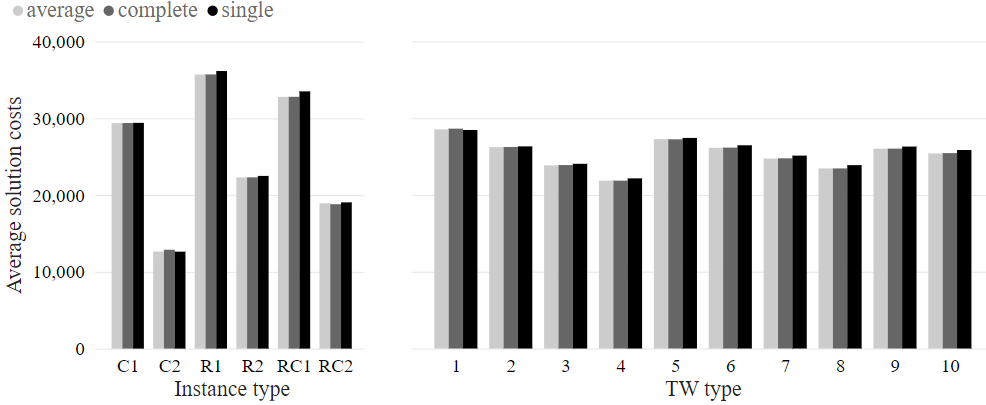}
    \end{adjustbox}
    \caption{Impact of linkage methods in agglomerative clustering on the routing costs}
    \label{fig:linkage_param}
\end{figure}

\begin{figure}[ht] 
    \centering
    \begin{adjustbox}{width=0.85\linewidth}
        \includegraphics[width=\linewidth]{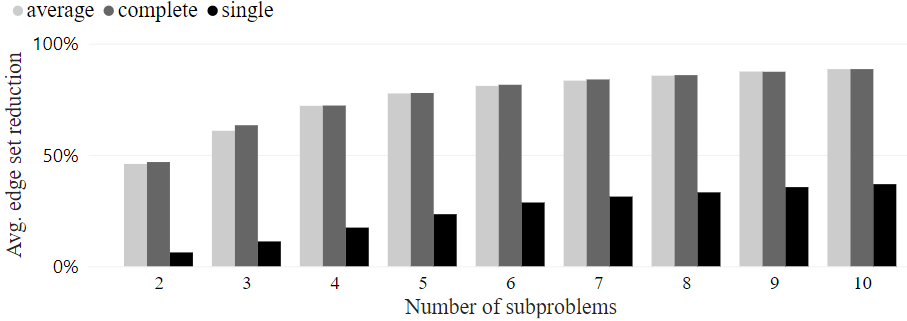}
    \end{adjustbox}
    \caption{Impact of linkage methods in agglomerative clustering on the reduction of the edge set}
    \label{fig:linkage_dim_red}
\end{figure}

We first analyze approach-specific parameters before we compare the clustering algorithms introduced in \ref{cluster_alg}. In fuzzy clustering, $\kappa$, the parameter that controls the fuzziness of the clusters, is usually set to $2$ \citep{Bezdek_81_fcm}. The higher $\kappa$, the more fuzzy the clusters are. Figure \ref{fig:fuzzy_m_param} shows that the solution costs of the \textit{decomposition} and \textit{routing phase} are not very sensitive to different values for $\kappa$ for different customer, vehicle capacity, or time window characteristics. That is a consequence of the design of the DRI, where we assign each customer to a single subproblem. In fuzzy c-medoids, a customer is assigned to that cluster where its degree of membership value is maximum.
In agglomerative clustering, the formation of the clusters depends on the linkage method. In Section \ref{cluster_alg}, we presented \textit{single}, \textit{complete}, and \textit{average} linkage. Figure \ref{fig:linkage_param} shows that \textit{single} linkage only works well for C instances where customers are in dense areas sparsely distributed around the depot. Thus, combining customers following the principle of the NN heuristic leads to high-quality sub-VRPTWs.
That is also true when the time windows are tight, i.e., TW type 1. This time window scenario makes the routes short and compact, as most routing options over a longer distance are infeasible. For all other instance and TW types, \textit{average} and \textit{complete} linkage result in lower solution costs.
The poor performance of the \textit{single} linkage is interesting, as it also fails to efficiently reduce the size of the edge set $E$ (i.e., distance matrix) as shown in Figure \ref{fig:linkage_dim_red}. The complexity reduction is calculated as the sum of the subproblems' edge sets size $|E_{p}|$ relative to the size of the edge set of the original problem $E$, i.e., ${|E|^{-1}\cdot \sum_{p=1}^q |E_{p}|}$. Theoretically, a less reduced distance matrix leads to more routing options and better route plans (when runtime is not a limiting factor). The more clusters are created, the higher the possible reduction of the edge set size. However, \textit{single} linkage tends to form one large subproblem of almost the same size as the original problem, while other subproblems only consist of a few customers. That is because of its greedy strategy that combines clusters solely based on the most similar customer pair, leading to an unfavorable structure of the subproblems. The linkage methods \textit{complete} and \textit{average} form evenly sized subproblems and thus achieve similar reduction scores. Thus, we disregard \textit{single} linkage in the following analyses.

\subsubsection{Impact of similarity metric in the decomposition phase:}
\begin{figure}[ht] 
    \centering
    \begin{adjustbox}{width=0.85\linewidth}
        \includegraphics[width=\linewidth]{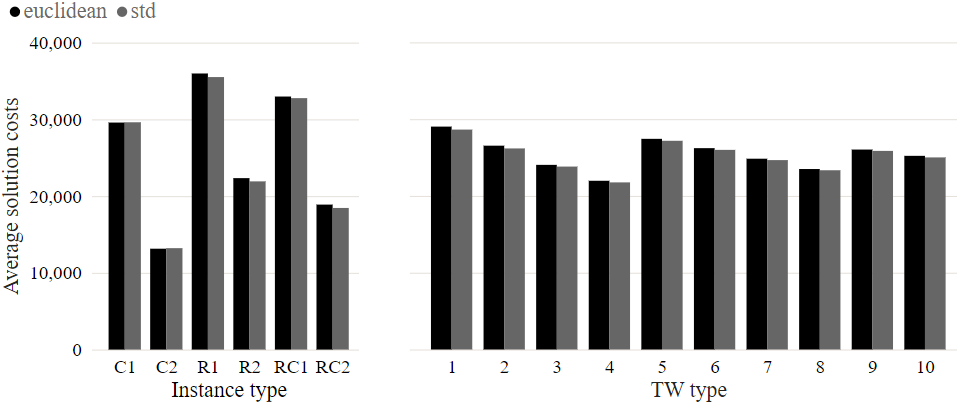}
    \end{adjustbox}
    \caption{Impact of the similarity metric ($\mathfrak{S}_{i,j}^{std}$) on the routing costs in comparison to the baseline ($C(e_{i,j})$)}
    \label{fig:dr_eucl_vs_std}
\end{figure}

Figure \ref{fig:dr_eucl_vs_std} compares the results when decomposing the original instance based on $\mathfrak{S}_{i,j}^{std}$ using the complete feature vector $\tau_{i}$ against the baseline metric $C(e_{i,j})$. This baseline is highly competitive, as, by definition, travel costs are the primary driver of solution quality.
Our results show that the STD metric is, on average, superior for all problem characteristics. Thus, it is beneficial to use spatial information in combination with temporal and demand data to measure customer similarity and effectively group customers into separate subsets in the \textit{decomposition phase}.
Table \ref{tab:dr_clust_meth} lists the solution of the best run of each clustering method and $q$ for instances that include $1000$ customers. We only report routing costs for time window scenarios $1, 10, 4$, and $6$ to improve readability. Bold values represent the best solution per row. For C instances, agglomerative clustering constantly yields the best solution, regardless of whether the routes are short (C1) or long (C2) or different time window scenarios are active. For C1 instances, solution costs increase slightly with increasing $q$. Thus, agglomerative clustering can efficiently separate C1 instances into multiple subproblems. Therefore, these instances show high scalability potential as their relatively short routes do not spread over the complete spatial plane.
\newpage
Thus, additional splits do not worsen the final solution quality even when the size of $E$ is reduced by up to 80\%, i.e., $q = 6$. For C2 instances, transportation costs slightly increase with the number of clusters. That is mainly because longer routes make it difficult to efficiently separate many subproblems. For R2 instances, k-medoids constantly achieves the best results as the size of the clusters is well-balanced. For R1 and RC1 instances, agglomerative clustering and k-medoids achieve similar results. Similar to C1 instances, more clusters can be created without downgrading the solution significantly. Among the three clustering algorithms, fuzzy c-medoids only finds the overall best solution once (instance RC2\_10\_6). Hence, deterministic clustering is superior in the context of decomposing routing problems. This result is not surprising as we have to allocate each customer to a single subproblem. The fuzzy information is particularly relevant for LS operations in the \textit{improvement phase}.
    \begin{table}
        \rotatebox{90}{
            \begin{minipage}{0.95\textheight}
                \small
                \centering
                \caption{\label{tab:dr_clust_meth} Solution costs of agglomerative clustering (ac), fuzzy c-medoids (fcm), and k-medoids (k-m) for different number of subproblems $q$}
                \begin{tabular}{r|ccc|ccc|ccc|ccc|ccc}
\toprule
q \ &     \multicolumn{3}{c|}{2} &     \multicolumn{3}{c|}{3} &     \multicolumn{3}{c|}{4} &     \multicolumn{3}{c|}{5} &     \multicolumn{3}{c}{6}\\ 
{} &     ac  & fcm  & k-m &     ac  & fcm  & k-m &     ac  & fcm  & k-m &     ac  & fcm  & k-m &     ac  & fcm  & k-m\\ 
\midrule 

C1\_10\_1 &  \textbf{42,479} & 42,876 & 42,783 &         \textbf{42,479} & 43,270 & \textbf{42,479} &         \textbf{42,479} & 42,550 & 42,481 &         42,482 & 43,236 & 42,499 &         42,487 & 43,351 & 42,753 \\ 
C1\_10\_10 &  40,200 & 40,621 & 40,438 &         40,210 & 40,873 & 40,365 &         40,174 & 40,711 & 40,369 &         40,115 & 41,044 & 40,494 &         \textbf{40,087} & 41,080 & 40,222 \\ 
C1\_10\_4 &  39,695 & 39,969 & 39,798 &         39,658 & 39,990 & 39,692 &         39,646 & 40,079 & 39,750 &         \textbf{39,642} & 40,278 & 39,940 &         39,734 & 40,397 & 39,912 \\ 
C1\_10\_6 &  \textbf{42,471} & 42,768 & \textbf{42,471} &         \textbf{42,471} & 43,541 & \textbf{42,471} &         \textbf{42,471} & 43,565 & 42,491 &         42,474 & 44,510 & 42,472 &         42,479 & 44,809 & 43,282 \\ 
\hline
C2\_10\_1 &  \textbf{16,879} & 17,603 & 17,277 &         16,950 & 18,429 & 17,629 &         16,973 & 18,875 & 17,703 &         16,973 & 18,338 & 18,104 &         17,034 & 18,193 & 18,542 \\ 
C2\_10\_10 &  \textbf{15,813} & 16,120 & 15,962 &         15,892 & 16,872 & 15,938 &         15,947 & 16,811 & 16,114 &         15,963 & 17,088 & 16,490 &         16,104 & 16,764 & 16,753 \\ 
C2\_10\_4 &  \textbf{15,589} & 15,774 & 15,690 &         15,672 & 16,407 & 15,704 &         15,702 & 16,283 & 15,623 &         15,724 & 16,742 & 15,978 &         15,846 & 16,195 & 16,430 \\ 
C2\_10\_6 &  \textbf{16,362} & 16,816 & 16,531 &         16,434 & 17,560 & 16,826 &         16,459 & 17,869 & 17,281 &         16,476 & 17,851 & 17,067 &         16,583 & 17,382 & 17,423 \\ 
\hline
R1\_10\_1 &  54,462 & 54,427 & 54,570 &         54,216 & 54,506 & 54,495 &         \textbf{54,207} & 54,536 & 54,220 &         54,350 & 54,811 & 54,736 &         54,879 & 55,713 & 55,242 \\ 
R1\_10\_10 &  48,450 & 48,337 & 48,427 &         48,514 & 48,368 & 48,403 &         48,512 & 48,405 & \textbf{48,215} &         48,814 & 48,343 & 48,221 &         48,934 & 49,398 & 48,450 \\ 
R1\_10\_4 &  43,137 & 43,158 & 43,049 &         43,050 & 43,111 & 43,142 &         \textbf{42,989} & 43,170 & 43,009 &         43,056 & 43,179 & 43,224 &         43,157 & 43,707 & 43,217 \\ 
R1\_10\_6 &  48,121 & 48,251 & 48,027 &         48,005 & 48,292 & \textbf{47,845} &         47,924 & 48,056 & 47,877 &         48,100 & 48,146 & 47,931 &         48,226 & 48,788 & 47,909 \\ 
\hline
R2\_10\_1 &  37,827 & 37,638 & \textbf{37,449} &         38,117 & 37,901 & 37,630 &         38,374 & 39,887 & 37,792 &         39,166 & 39,775 & 38,259 &         39,661 & 40,494 & 38,604 \\ 
R2\_10\_10 &  31,049 & 30,774 & \textbf{30,769} &         31,231 & 30,881 & 30,940 &         31,395 & 32,582 & 30,943 &         32,166 & 32,571 & 31,590 &         32,583 & 32,732 & 33,004 \\ 
R2\_10\_4 &  18,603 & 18,461 & \textbf{18,245} &         18,634 & 18,330 & 18,359 &         18,756 & 19,629 & 18,319 &         19,153 & 19,459 & 18,663 &         19,572 & 19,894 & 18,858 \\ 
R2\_10\_6 &  30,321 & 30,166 & \textbf{29,818} &         30,401 & 30,124 & 30,014 &         30,597 & 31,873 & 29,970 &         31,395 & 31,679 & 30,303 &         31,845 & 32,451 & 30,701 \\ 
\hline
RC1\_10\_1 &  46,597 & 46,669 & 46,629 &         46,581 & 46,839 & 46,690 &         \textbf{46,571} & 46,620 & 46,585 &         46,729 & 46,654 & 46,733 &         46,740 & 47,073 & 47,004 \\ 
RC1\_10\_10 &  44,105 & 44,438 & \textbf{44,078} &         44,129 & 44,512 & 44,259 &         44,257 & 44,274 & 44,246 &         44,249 & 44,632 & 44,311 &         44,316 & 44,972 & 44,422 \\ 
RC1\_10\_4 &  41,920 & 42,018 & 41,866 &         \textbf{41,786} & 42,076 & 41,960 &         41,852 & 42,124 & 42,042 &         41,873 & 42,188 & 42,178 &         41,815 & 42,679 & 42,138 \\ 
RC1\_10\_6 &  \textbf{45,596} & 46,028 & 45,689 &         45,657 & 46,076 & 45,673 &         45,801 & 45,863 & 45,698 &         45,782 & 46,247 & 45,690 &         45,858 & 46,573 & 45,951 \\ 
\hline
RC2\_10\_1 &  \textbf{28,527} & 28,725 & 28,747 &         29,038 & 29,734 & 28,635 &         29,369 & 29,616 & 29,000 &         29,648 & 30,576 & 29,194 &         30,081 & 30,970 & 29,558 \\ 
RC2\_10\_10 &  22,306 & 22,307 & \textbf{22,248} &         22,619 & 23,285 & 22,421 &         22,765 & 23,057 & 22,855 &         22,930 & 23,729 & 23,147 &         23,255 & 24,227 & 24,092 \\ 
RC2\_10\_4 &  \textbf{15,985} & 16,008 & 16,014 &         16,259 & 16,876 & 16,164 &         16,538 & 16,630 & 16,361 &         16,655 & 17,327 & 16,537 &         16,839 & 17,725 & 16,960 \\ 
RC2\_10\_6 &  26,397 & \textbf{26,372} & 26,387 &         26,794 & 27,504 & 26,420 &         26,982 & 27,273 & 26,608 &         27,160 & 28,107 & 26,896 &         27,521 & 28,493 & 27,094 \\ 
\bottomrule
\end{tabular}

            \end{minipage}
        }
    \end{table}
\subsection{Hyperparameters - improvement phase}
The impact of the \textit{improvement phase} depends on the clusters created by the \textit{decomposition phase} and their solutions found by the \textit{routing phase}. The fewer subproblems and the more runtime allocated to them, the better their solutions, making it harder to find improvements when combining the separate route plans. However, for an autonomous evaluation of the impact of hyperparameters in the \textit{improvement phase}, the solutions of the subproblems yielded by the \textit{routing phase} should include unfavorable routing decisions for the overall solution. This is achieved by splitting instances into many subproblems and setting short runtimes for solving. Thus, we consider instances with $1000$ customers, set $q = 10$ in the \textit{decomposition phase}, and $\Omega = 60$ seconds in the \textit{routing phase}. This computation time is shared between the subproblems proportionally to their size, i.e., a cluster $p$ with $100$ customers must be solved in $\Omega_{p} = 6$ seconds, and the stopping criterion $\Omega_{g}$ for a cluster $g$ with, e.g., $120$, is $7$ seconds.
\subsubsection{Internal parameters:}
We calculate the relative difference of the error gap $\xi_\mathfrak{R_{I}}$ ($\xi_\mathfrak{R_{I}}'$) before (after) the \textit{improvement phase} in (\ref{eq:rel_diff_e_gap_ls}). The \textit{error gap} (\ref{eq:e_gap}) is defined as the relative difference between the total cost of the solution generated $Z_{\mathfrak{R_{I}}}$ and the total costs of the best known solution (BKS) $Z_{\mathfrak{R_{I}}^*}$. BKS solutions of all benchmark instances are obtained from the CVRPLIB website at \emph{http://vrp.atd-lab.inf.puc-rio.br/index.php/en/}(Accessed on October 1st, 2023).
\begin{equation}
    \label{eq:rel_diff_e_gap_ls}
    \widetilde{\xi} = \frac{\xi_\mathfrak{R_{I}}' - \xi_\mathfrak{R_{I}}}{\xi_\mathfrak{R_{I}}}
\end{equation}
\begin{equation}
    \label{eq:e_gap}
    \xi_{\mathfrak{R_{I}}} = \frac{
        Z_{\mathfrak{R_{I}}} - Z_{\mathfrak{R_{I}}^*}}{
        Z_{\mathfrak{R_{I}}^*}
    }
\end{equation}
The effectiveness (i.e., achieved improvement of the overall solution $\mathfrak{R_{I}}$, and the runtime until a local optimum is reached) is mainly driven by the pruning factors $\phi$, i.e., the number of route-pairs available for LS, and $\varphi$, i.e., the number of vertices available for a specific LS move.
The parameter $\phi$ defines the number of subproblems in $\Phi_{p}$. Thus, $\phi$ limits the number of route-pairs considered in LS. Figure \ref{fig:ls_subproblem_vicinity_e_gap} and Figure \ref{fig:ls_subproblem_vicinity_rt} report the impact of ${\phi = \{1, 3, 5, 7, 9\}}$ on the LS efficiency. For $\phi = 9$, no route-pairs are excluded as $q = 10$. With increasing $\phi$, the overall solution is improving on average. This effect reaches a plateau for $\phi = 5$. Also, the runtime increases as LS moves are applied on more route-pairs but without leading to noteworthy further improvements.  
\begin{minipage}{\linewidth}
      \centering
      \begin{minipage}{0.3\linewidth}
          \begin{figure}[H]
              \includegraphics[width=\linewidth]{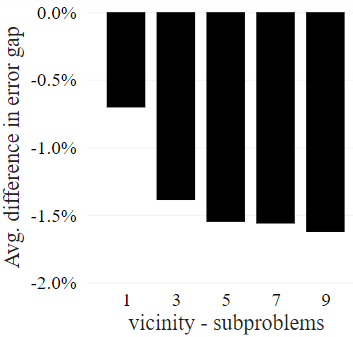}
              \caption{Impact of $\Phi$ on error gap}
              \label{fig:ls_subproblem_vicinity_e_gap}
          \end{figure}
      \end{minipage}
      \hspace{0.15\linewidth}
      \begin{minipage}{0.3\linewidth}
          \begin{figure}[H]
              \includegraphics[width=\linewidth]{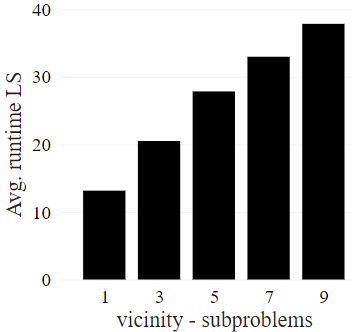}
              \caption{Impact of $\Phi$ on LS runtime}
              \label{fig:ls_subproblem_vicinity_rt}
          \end{figure}
      \end{minipage}
\end{minipage}
The parameter $\varphi$ sets the size of the vicinity $\Phi_{i}$ on the vertex level. Thus, it limits the number of moves possible in a LS operation. Figure \ref{fig:ls_vertex_vicinity_e_gap} reports the average $\widetilde{\xi}$ and Figure \ref{fig:ls_vertex_vicinity_rt} shows the average runtime of the \textit{improvement phase} for ${\varphi = \{5, 10, 30, 1000\}}$. If $\varphi = 1000$, no pruning is active. When the goal solely is to compensate for potentially inefficient routing decisions along the perimeters of the subproblems fast, on average, it is sufficient to limit LS moves to the $10$ most similar vertices. For smaller $\varphi$, a local optimum is reached faster.
\begin{minipage}{\linewidth}
      \centering
      \begin{minipage}{0.28\linewidth}
          \begin{figure}[H]
              \includegraphics[width=\linewidth]{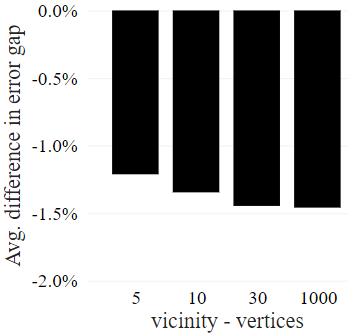}
              \caption{Impact of $\varphi$ on error gap}
              \label{fig:ls_vertex_vicinity_e_gap}
          \end{figure}
      \end{minipage}
      \hspace{0.15\linewidth}
      \begin{minipage}{0.28\linewidth}
          \begin{figure}[H]
\includegraphics[width=\linewidth]{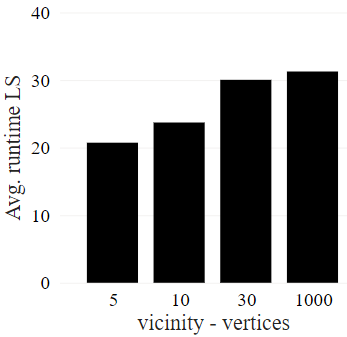}
              \caption{Impact of $\varphi$ on LS runtime}
              \label{fig:ls_vertex_vicinity_rt}
          \end{figure}
      \end{minipage}
\end{minipage}
Table \ref{tab:ls_strategy_efficiency} lists the average reduction in error gap and average computation time for the two LS strategies \textit{first-descent} and \textit{steepest-descent} for different customer location distributions. \textit{Steepest-descent} is superior in improving a solution, particularly when customers are randomly located, i.e., R and RC. Thus, evaluating all available moves of a LS operator reduces the risk of getting stuck in a local optimum too early. The more moves are executed, the longer LS runs. However, on average, computation time of the \textit{steepest-descent} increases not proportionally with average solution improvement.
\begin{table}[htbp]
    \small
    \centering
    \caption{\label{tab:ls_strategy_efficiency} Impact of search strategy on LS efficiency}
        \begin{tabular}{c|cc|cc}
    \toprule
    customer & \multicolumn{2}{c|}{first-descent} & \multicolumn{2}{c}{steepest-descent}\\ 
    location & Avg. & Avg.  & Avg.  & Avg.\\ 
    distribution & $\widetilde{\xi}$ & $\Upsilon$ & $\widetilde{\xi}$ & $\Upsilon$\\
    \midrule 
    C   &   -0.74\%     & 38.54 & -0.86\% & 36.76 \\
    R   &   -0.31\%     & 41.56 & -1.14\% & 67.08 \\
    RC  &   -1.51\%     & 69.95 & -2.73\% & 78.28 \\    
    \bottomrule
\end{tabular}

\end{table}
\subsubsection{Impact of data-based approach in improvement phase:}
\begin{figure}[ht] 
    \centering
    \begin{adjustbox}{width=0.62\linewidth}
        \includegraphics[width=\linewidth]{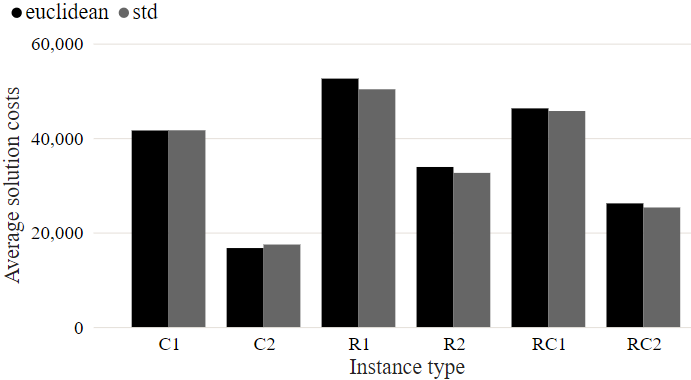}
    \end{adjustbox}
    \caption{Impact of the of the similarity metric $\mathfrak{S}_{i,j}^{std}$ on the LS efficiency - routing costs}
    \label{fig:ls_std_sol_costs}
\end{figure}

\begin{figure}
    \centering
    \begin{adjustbox}{width=0.62\linewidth}
        \includegraphics[width=\linewidth]{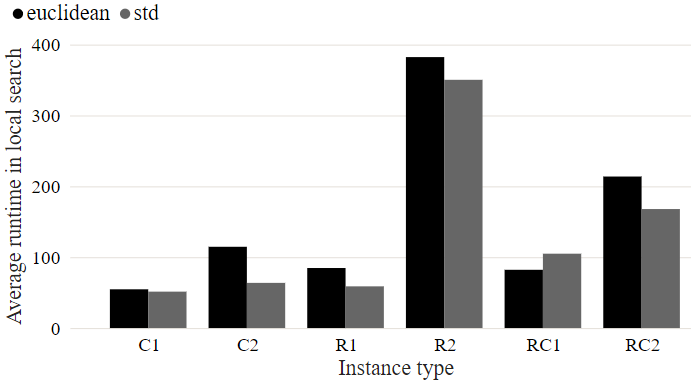}
    \end{adjustbox}
    \caption{Impact of the of the similarity metric $\mathfrak{S}_{i,j}^{std}$ on the LS efficiency - runtime}
    \label{fig:ls_std_rt}
\end{figure}

Our data-based approach aims to guide LS in the \textit{improvement phase}. To achieve efficient pruning, it is essential to carefully select the vertices of $\Phi_{i}$. Typically, one would select the most promising vertices to the solution, i.e., based on $C(e_{i,j})$. However, Figure \ref{fig:ls_std_sol_costs} and Figure \ref{fig:ls_std_rt} show that using the STD distance that includes the penalization based on feasibility improves the efficiency of LS in comparison to the Euclidean distance between the customer coordinates. When fuzzy c-medoids is used in the \textit{decomposition phase}, we receive additional information about customers' degree of membership to subproblems. Fuzzy vertices are located along the boundaries of neighboring clusters. A customer $i$ is labeled fuzzy when $\argmax_{V_{p}}(\mu_{i, V_{p}}) < \rho$. For $\rho < 1$, LS moves are pruned exclusively to such fuzzy customers.

Figure \ref{fig:ls_fuzzy_e_gap} shows improvements even when the threshold is strict, i.e., $\rho \leq 0.5$. The more customers are marked fuzzy, i.e., the larger $\rho$, the more LS operations are executed and the better the final solution. However, Figure \ref{fig:ls_fuzzy_rt} shows that the runtime grows exponentially with increasing $\rho$, indicating that most modifications appear near neighboring subproblems' perimeters.  Thus, data-based LS that leverages information obtained from the \textit{decomposition phase} is particularly beneficial when the available computation time for the \textit{improvement phase} is short.
\begin{minipage}{\linewidth}
      \centering
      \begin{minipage}{0.31\linewidth}
          \begin{figure}[H]
              \includegraphics[width=\linewidth]{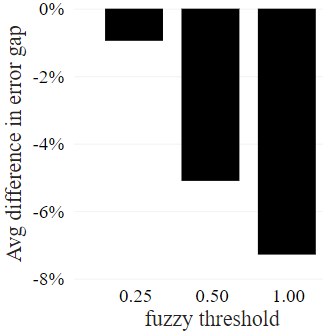}
              \caption{Impact of $\rho$ on error gap improvement}
              \label{fig:ls_fuzzy_e_gap}
          \end{figure}
      \end{minipage}
      \hspace{0.15\linewidth}
      \begin{minipage}{0.31\linewidth}
          \begin{figure}[H]
              \includegraphics[width=\linewidth]{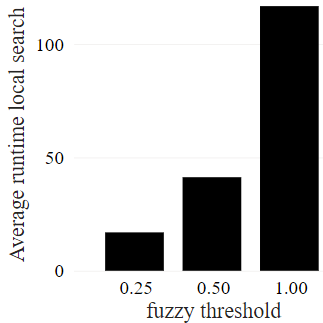}
              \caption{Impact of $\rho$ on LS runtime}
              \label{fig:ls_fuzzy_rt}
          \end{figure}
      \end{minipage}
\end{minipage}
\subsubsection{Runtime budgeting in DRI:}
\begin{figure}[htbp] 
    \centering
    \begin{adjustbox}{width=0.82\linewidth}
        \includegraphics[width=\linewidth]{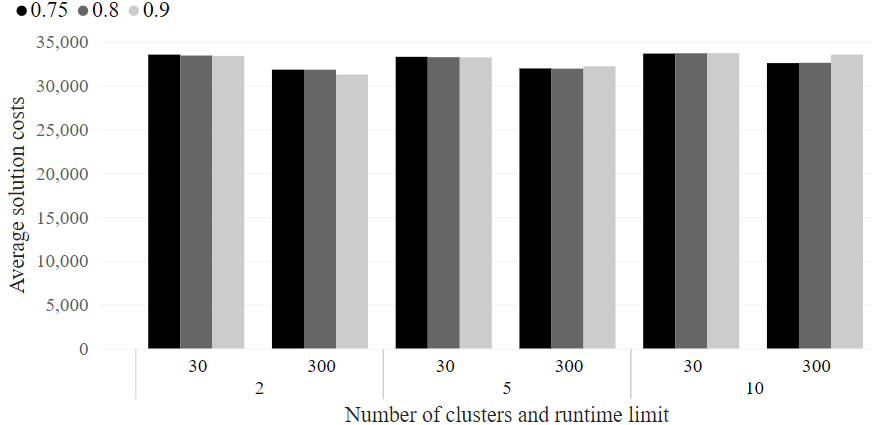}
    \end{adjustbox}
    \caption{Impact of the runtime allocation $\alpha$ on the routing costs}
    \label{fig:time_budget}
\end{figure}

As introduced in Section \ref{sol_frame:route}, the computation time must be allocated between the three phases of the DRI, i.e., \textit{decomposition}, \textit{routing}, and \textit{improvement}. Even for large instances with $1000$ customers, the clustering barely consumes any runtime, i.e., $\nu \leq 0.5$ seconds. Thus, we focus on the runtime split ${\alpha =\{0.75, 0.8, 0.9\}}$ of the available time between the \textit{routing} ($\alpha$) and \textit{improvement phase} ($1- \alpha$). LS aims to revise the solution found in the previous steps. Thus, we allocate most computation time to the \textit{routing phase}.
The hyperparameters are set based on the analyses presented in the previous sections and are listed in Table \ref{tab:time_budget_dri_hyper}. Two total runtime limits are applied: ${\Delta = \{30, 300\}}$.
Figure \ref{fig:time_budget} shows that the shorter the total runtime and the fewer clusters are created, the more runtime should be allocated to the \textit{routing phase}. Here, spending more time to solve the subproblems allows us to find better solutions than running an exhaustive LS between the subproblems. With increasing runtime and more subproblems created, reserving more time for the \textit{improvement phase} is beneficial. Then, the subproblems are likely to be solved close to optimality, and more improvements can be found in the regions where the customers have been split.
\begin{table}[htbp]
    \footnotesize
    \centering
    \begin{threeparttable}
        \caption{\label{tab:time_budget_dri_hyper} DRI hyperparameter setup for computational time budgeting analysis}
        \begin{tabular}{r|c}
\toprule
hyperparameter  & values \\ 
\midrule 
number of clusters  & $\{2, 5, 10\}$ \\
$\theta_{i}$        & $1$ \\
clustering method   & k-medoids, agglomerative clustering \\
similarity metric   & $\mathfrak{S}_{i,j}^{std}$ \\
\midrule
$\phi$              & $5$ \\
$\varphi$           & $10$ \\
search strategy     & \textit{steepest-descent} \\
\bottomrule
\end{tabular}
    \end{threeparttable}
\end{table}
\newpage
\subsubsection{Comparison against state-of-the-art solver:}
We compare the DRI framework with the standard HGS-TW implementation (https://pyvrp.org/) to evaluate its potential to increase scalability of state-of-the-art solution methods. Therefore, we compare the quality of the solution of the two algorithms after specific runtimes. Here, the focus lies on short computation times, as our goal is to develop a solution that finds solutions of high-quality fast. Thus, we set ${\Theta = \{15, 30, 45, 60, 75, 90, 120, 300\}}$. Table \ref{tab:dri_hgs_tw_overview} lists the average of the best solutions found for different instance sizes per instance and time window type. The bold values highlight the best solution per runtime limit and instance. Tables \ref{tab:dri_hgs_tw_600}, \ref{tab:dri_hgs_tw_800}, and \ref{tab:dri_hgs_tw_1000} in the Appendix list the best solutions found for the instances with $600, 800$, $1000$ customers respectively. On average, the DRI outperforms the HGS-TW for most instances for $\Theta \leq 60$. When routes are short (C1, R1, RC1), splitting instances into multiple smaller subproblems, routing them separately, and applying data-based LS outperforms the standard HGS-TW over all tested runtimes. Table \ref{tab:dri_hgs_tw_600} shows that for C instances with $600$ customers, the HGS-TW converges within $60$ seconds. Thus, decomposition does not improve scalability for this instance type for medium-sized problems. However, the DRI finds better solutions for the R and RC instances for short runtimes, i.e., $\Theta = \{15, 30\}$. With increasing problem sizes (Tables \ref{tab:dri_hgs_tw_1000} and \ref{tab:dri_hgs_tw_800}), the DRI reveals its strength in scaling state-of-the-art solution algorithms. 
\begin{table}
    \small
    \rotatebox{90}{
        \begin{minipage}{0.95\textheight}
    \centering
        \caption{\label{tab:dri_hgs_tw_overview} Comparison of solution quality between DRI and HGS-TW for different runtimes}
        \begin{tabular}{rr|cc|cc|cc|cc|cc|cc|cc|cc}
\toprule
{} & $\Theta$   &     \multicolumn{2}{c|}{15} &     \multicolumn{2}{c|}{30} &     \multicolumn{2}{c|}{45} &     \multicolumn{2}{c|}{60} &     \multicolumn{2}{c|}{75} &     \multicolumn{2}{c|}{90} &     \multicolumn{2}{c|}{120} &     \multicolumn{2}{c}{300} \\ 
$\mathfrak{I}$-type & TW &     DRI  & HGS &     DRI  & HGS &     DRI  & HGS &     DRI  & HGS &     DRI  & HGS &     DRI  & HGS &     DRI  & HGS &     DRI  & HGS \\ 
\midrule 
C1 & 1 &         27,428 & \textbf{27,253}   &         27,428 & \textbf{27,253}   &         27,428 & \textbf{27,253}   &         27,428 & \textbf{27,253}   &         27,428 & \textbf{27,253}   &         27,428 & \textbf{27,253}   &         27,428 & \textbf{27,253} &         27,428 & \textbf{27,253} \\ 
C1 & 10 &         \textbf{26,733} & 27,251   &         \textbf{26,478} & 26,846   &         \textbf{26,446} & 26,648   &         \textbf{26,390} & 26,648   &         \textbf{26,363} & 26,545   &         \textbf{26,371} & 26,504   &         \textbf{26,289} & 26,437 &         \textbf{26,139} & 26,281 \\ 
C1 & 4 &         \textbf{26,373} & 27,123   &         \textbf{26,205} & 26,650   &         \textbf{26,191} & 26,519   &         \textbf{26,124} & 26,400   &         \textbf{26,043} & 26,337   &         \textbf{26,026} & 26,326   &         \textbf{25,971} & 26,313 &         \textbf{25,824} & 25,927 \\ 
C1 & 6 &         27,351 & \textbf{27,250}   &         27,351 & \textbf{27,241}   &         27,351 & \textbf{27,240}   &         27,351 & \textbf{27,240}   &         27,351 & \textbf{27,240}   &         27,351 & \textbf{27,240}   &         27,351 & \textbf{27,240} &         27,351 & \textbf{27,240} \\ 
\midrule 
C2 & 1 &         \textbf{12,722} & 13,053   &         12,577 & \textbf{12,282}   &         12,575 & \textbf{12,256}   &         12,575 & \textbf{12,105}   &         12,575 & \textbf{12,105}   &         12,575 & \textbf{12,105}   &         12,575 & \textbf{12,105} &         12,575 & \textbf{12,105} \\ 
C2 & 10 &         \textbf{12,090} & 13,112   &         \textbf{11,692} & 12,150   &         \textbf{11,647} & 11,708   &         11,602 & \textbf{11,550}   &         11,570 & \textbf{11,431}   &         11,557 & \textbf{11,402}   &         11,546 & \textbf{11,347} &         11,514 & \textbf{11,324} \\ 
C2 & 4 &         \textbf{11,882} & 12,416   &         \textbf{11,392} & 11,869   &         \textbf{11,367} & 11,577   &         \textbf{11,330} & 11,460   &         11,275 & \textbf{11,256}   &         11,301 & \textbf{11,205}   &         11,262 & \textbf{11,105} &         11,220 & \textbf{11,056} \\ 
C2 & 6 &         \textbf{12,519} & 13,048   &         12,102 & \textbf{11,943}   &         12,080 & \textbf{11,804}   &         12,062 & \textbf{11,747}   &         12,045 & \textbf{11,740}   &         12,050 & \textbf{11,733}   &         12,041 & \textbf{11,723} &         12,038 & \textbf{11,718} \\ 
\midrule 
R1 & 1 &         \textbf{39,371} & 117,926   &         \textbf{38,645} & 117,441   &         \textbf{38,410} & 117,165   &         \textbf{38,326} & 117,078   &         \textbf{38,252} & 38,552   &         \textbf{38,133} & 38,342   &         \textbf{38,069} & 38,115 &         37,781 & \textbf{37,759} \\ 
R1 & 10 &         \textbf{33,756} & 33,907   &         \textbf{33,452} & 33,746   &         \textbf{33,399} & 33,521   &         \textbf{33,239} & 33,350   &         \textbf{33,077} & 33,208   &         \textbf{33,006} & 33,128   &         \textbf{32,852} & 33,036 &         \textbf{32,559} & 32,757 \\ 
R1 & 4 &         \textbf{30,074} & 30,462   &         \textbf{29,920} & 30,109   &         \textbf{29,871} & 29,962   &         \textbf{29,791} & 29,905   &         \textbf{29,757} & 29,833   &         \textbf{29,676} & 29,782   &         \textbf{29,496} & 29,672 &         \textbf{29,135} & 29,262 \\ 
R1 & 6 &         \textbf{33,810} & 34,073   &         \textbf{33,616} & 33,823   &         \textbf{33,474} & 33,676   &         \textbf{33,345} & 33,561   &         \textbf{33,327} & 33,494   &         \textbf{33,239} & 33,385   &         \textbf{33,105} & 33,204 &         \textbf{32,696} & 32,878 \\ 
\midrule 
R2 & 1 &         \textbf{27,363} & 28,376   &         \textbf{26,811} & 27,216   &         \textbf{26,598} & 26,716   &         26,544 & \textbf{26,512}   &         26,462 & \textbf{26,318}   &         26,356 & \textbf{26,284}   &         26,285 & \textbf{26,145} &         26,064 & \textbf{25,881} \\ 
R2 & 10 &         \textbf{22,382} & 22,578   &         \textbf{21,746} & 21,878   &         21,679 & \textbf{21,677}   &         \textbf{21,505} & 21,552   &         \textbf{21,393} & 21,429   &         \textbf{21,328} & 21,342   &         21,291 & \textbf{21,223} &         21,128 & \textbf{20,883} \\ 
R2 & 4 &         \textbf{13,973} & 14,284   &         \textbf{13,716} & 13,978   &         \textbf{13,673} & 13,784   &         \textbf{13,608} & 13,675   &         \textbf{13,479} & 13,640   &         \textbf{13,465} & 13,585   &         \textbf{13,381} & 13,497 &         13,289 & \textbf{13,248} \\ 
R2 & 6 &         \textbf{22,051} & 22,446   &         \textbf{21,515} & 21,913   &         \textbf{21,392} & 21,658   &         \textbf{21,280} & 21,465   &         \textbf{21,176} & 21,329   &         \textbf{21,150} & 21,208   &         \textbf{21,017} & 21,109 &         \textbf{20,801} & 20,846 \\ 
\midrule 
RC1 & 1 &         \textbf{32,525} & 32,862   &         \textbf{32,232} & 32,593   &         \textbf{32,077} & 32,373   &         \textbf{31,957} & 32,272   &         \textbf{31,942} & 32,094   &         \textbf{31,846} & 32,050   &         \textbf{31,807} & 31,890 &         \textbf{31,531} & 31,610 \\ 
RC1 & 10 &         \textbf{30,498} & 30,852   &         \textbf{30,231} & 30,498   &         \textbf{30,114} & 30,341   &         \textbf{30,052} & 30,246   &         \textbf{30,012} & 30,177   &         \textbf{29,920} & 30,115   &         \textbf{29,801} & 29,952 &         \textbf{29,560} & 29,660 \\ 
RC1 & 4 &         \textbf{28,782} & 28,982   &         \textbf{28,529} & 28,820   &         \textbf{28,504} & 28,697   &         \textbf{28,447} & 28,617   &         \textbf{28,380} & 28,535   &         \textbf{28,349} & 28,499   &         \textbf{28,197} & 28,397 &         \textbf{28,034} & 28,098 \\ 
RC1 & 6 &         \textbf{31,799} & 32,058   &         \textbf{31,423} & 31,645   &         \textbf{31,382} & 31,497   &         \textbf{31,237} & 31,370   &         \textbf{31,145} & 31,229   &         \textbf{31,052} & 31,220   &         \textbf{30,969} & 31,123 &         \textbf{30,684} & 30,819 \\ 
\midrule 
RC2 & 1 &         \textbf{21,217} & 22,686   &         \textbf{20,642} & 21,401   &         \textbf{20,527} & 20,900   &         \textbf{20,409} & 20,700   &         \textbf{20,364} & 20,512   &         \textbf{20,326} & 20,335   &         20,301 & \textbf{20,150} &         20,178 & \textbf{19,918} \\ 
RC2 & 10 &         \textbf{16,311} & 16,540   &         \textbf{15,810} & 15,986   &         15,760 & \textbf{15,727}   &         15,731 & \textbf{15,651}   &         15,619 & \textbf{15,532}   &         15,586 & \textbf{15,494}   &         15,514 & \textbf{15,425} &         15,382 & \textbf{15,223} \\ 
RC2 & 4 &         \textbf{12,119} & 12,490   &         \textbf{11,831} & 12,207   &         \textbf{11,746} & 11,899   &         \textbf{11,715} & 11,781   &         \textbf{11,668} & 11,711   &         \textbf{11,638} & 11,659   &         11,605 & \textbf{11,560} &         11,491 & \textbf{11,376} \\ 
RC2 & 6 &         \textbf{19,501} & 20,363   &         \textbf{18,803} & 19,529   &         \textbf{18,661} & 19,019   &         \textbf{18,612} & 18,730   &         \textbf{18,527} & 18,575   &         \textbf{18,486} & 18,505   &         18,404 & \textbf{18,335} &         18,308 & \textbf{18,110} \\ 
\midrule 
\bottomrule
\end{tabular}

    \end{minipage}
}
\end{table}
\section{Conclusion}\label{cnclsn}
We developed the DRI to efficiently solve large routing problems. It combines data-based decomposition and pruning strategies with state-of-the-art routing methods. The computational study demonstrates that our proposed similarity metric, which includes customers' spatial, temporal, and demand information and reflects the problem's objective function and constraints, outperforms classic clustering metrics solely based on customer locations. Pruning is more efficient when the STD metric limits LS moves compared to a metric based only on travel costs. Our approach effectively addresses scalability issues of common heuristics.
\newpage
In the DRI, the more customers served, the more subproblems are formed, and the computation effort is shifted towards the \textit{improvement phase}. The edge set is most reducible, i.e., the DRI forms more subproblems, when customers are located closely together in sparsely distributed areas and routes are short. Fewer subproblems are required if the customers' distribution pattern is random and routes are long. Thus, if a feasible, high-quality solution needs to be generated fast for large-scale routing problems, the DRI outperforms current state-of-the-art solution algorithms. 
In the current setup, the LS only accepts strict improvements. Thus, it is not possible to escape local optima. Metaheuristics, i.e., simulated annealing or TS, can be used for complete improvement. As our goal is to focus on explicit route improvements, we refrain from implementing more complex search strategies in the improvement phase.
The cost function is solely calculated by the Euclidean distance between two vertices in the benchmark datasets. Investigating how the DRI performs on more complex cost functions that (i) consider the total driving time of the vehicles, i.e., also includes the waiting and service time at a customer location, and (ii) retrieves real-world travel data for a more sophisticated cost evaluation could yield valuable insights. Further, we plan to extend the DRI to other problem attributes, e.g., a heterogeneous fleet, multiple depots, and pickup and delivery requests. 

\section*{Acknowledgement}
The authors thank Anne Kißler, Roger Kowalewski, Marilena Leichter, and Heiner Zille from SAP SE for their helpful insights during the project. Moreover, the authors would like to thank Elif Erden and Jan Eichhorn for their help in coding the LS library. The research of Christoph Kerscher and Stefan Minner has been supported by SAP SE.

\newpage
\section*{Appendix}
    \begin{algorithm}[H]
        \caption{$k$-medoids}
        \label{alg:kmedoids}
        \begin{algorithmic}[1]
            \item Initialization: Select $q$ customer vertices following (\ref{eq:k_medoids_seed_calc}) as initial medoids.
            \Repeat
            \State{Assign each vertex $i$ to the closest cluster $V_{p^*}$, where ${p^* = \argmin_{p}( \overline{\mathfrak{S}}_{i,m_p}^{std})}$.}
            \State{Set customer $i$ that is most similar to all customers of that cluster as new medoid following (\ref{eq:k_medoids_m_p_update}).}
            \Until{$\{m_p, p=1,\ldots,q\}_t = \{m_p, p=1,\ldots,q\}_{t+1}$}
            \State{\textbf{return} set of customer clusters $\{V_{p}\}, \enspace p =1,\ldots,q$}
        \end{algorithmic}
    \end{algorithm}
    \begin{algorithm}[H]
        \caption{fuzzy c-medoids}
        \label{alg:f_c_m}
        \begin{algorithmic}[1]
            \State Initialization: $U^0$
            \While{$U^r - U^{r-1} < \epsilon $}\Comment{Convergence of the degrees of membership}\vskip 4pt
                \State{$\tau_{p} = \sum_{i=1}^{n} \mu_{i,V_{p}} \cdot \tau_i, \enspace p = 1,\ldots,q$}\vskip 4pt
                \State{$m_p=\argmin_{i \in V_{c}}(\overline{\mathfrak{S}}_{i,p}^{std}), \enspace p=1,\ldots,q$
                }\vskip 4pt
                \State{updated
                    $\mu_{i,V_{p}} \enspace \forall \enspace i \in V_{c}, \enspace p=1,\ldots,q$ following (\ref{eq:fcm_degree_of_membership})
                }\vskip 4pt
            \EndWhile\label{while_fcm}
            \State{\textbf{return} $U^r$, $\{m_p, p=1,\ldots,q\}$}
        \end{algorithmic}
    \end{algorithm}
    \begin{algorithm}[H]
        \caption{agglomerative clustering}
        \label{alg:ac}
        \begin{algorithmic}[1]
            \item Let $V_{p} = {i} \enspace \forall \enspace i = 1,\ldots,n$ and let $\mathfrak{V}=\{V_{p}\}$ where $p = 1,\ldots,n$
            \While{$|\mathfrak{V}| \geq q$}
                \State{Select cluster-pair ($V_{p}$, $V_{g}$) based on (\ref{eq:of_agglom})}
                \State{$V_{p^*} = V_{p} \cup V_{g}$}
                \State{$\mathfrak{V^*}= \mathfrak{V} \cup V_{p^*}\setminus \{V_{p},V_{g}$\}}
            \EndWhile
            \State{\textbf{return} set of customer clusters $\mathfrak{V^*}$}
        \end{algorithmic}
    \end{algorithm}

\begin{landscape}
\begin{table} [ht]
    \centering
    \small
    \begin{threeparttable}
        \caption{\label{tab:dri_hgs_tw_600} Comparison of solution quality between DRI and HGS-TW for different runtimes - instance size: 600}
        \begin{tabular}{r|cc|cc|cc|cc|cc|cc|cc|cc}
\toprule
$\Theta$   &     \multicolumn{2}{c|}{15} &     \multicolumn{2}{c|}{30} &     \multicolumn{2}{c|}{45} &     \multicolumn{2}{c|}{60} &     \multicolumn{2}{c|}{75} &     \multicolumn{2}{c|}{90} &     \multicolumn{2}{c|}{120} &     \multicolumn{2}{c}{300} \\ 
{} &     DRI  & HGS &     DRI  & HGS &     DRI  & HGS &     DRI  & HGS &     DRI  & HGS &     DRI  & HGS &     DRI  & HGS &     DRI  & HGS \\ 
\midrule 
C1\_6\_1      &         14,600 & \textbf{14,095}   &         14,600 & \textbf{14,095}   &         14,600 & \textbf{14,095}   &         14,600 & \textbf{14,095}   &         14,600 & \textbf{14,095}   &         14,600 & \textbf{14,095}   &         14,600 & \textbf{14,095} &         14,600 & \textbf{14,095} \\ 
C1\_6\_10      &         14,103 & \textbf{14,003}   &         13,970 & \textbf{13,772}   &         13,963 & \textbf{13,772}   &         13,964 & \textbf{13,772}   &         13,943 & \textbf{13,714}   &         13,939 & \textbf{13,713}   &         13,937 & \textbf{13,710} &         13,919 & \textbf{13,686} \\ 
C1\_6\_4      &         \textbf{13,989} & 14,129   &         13,905 & \textbf{13,873}   &         13,867 & \textbf{13,703}   &         13,838 & \textbf{13,676}   &         13,827 & \textbf{13,676}   &         13,811 & \textbf{13,674}   &         13,801 & \textbf{13,660} &         13,752 & \textbf{13,618} \\ 
C1\_6\_6      &         14,399 & \textbf{14,089}   &         14,399 & \textbf{14,089}   &         14,399 & \textbf{14,089}   &         14,400 & \textbf{14,089}   &         14,399 & \textbf{14,089}   &         14,399 & \textbf{14,089}   &         14,399 & \textbf{14,089} &         14,399 & \textbf{14,089} \\ 
\midrule 
C2\_6\_1      &         8,266 & \textbf{8,029}   &         8,241 & \textbf{7,776}   &         8,241 & \textbf{7,776}   &         8,241 & \textbf{7,776}   &         8,241 & \textbf{7,776}   &         8,241 & \textbf{7,776}   &         8,241 & \textbf{7,776} &         8,239 & \textbf{7,774} \\ 
C2\_6\_10      &         \textbf{7,573} & 7,593   &         7,480 & \textbf{7,229}   &         7,482 & \textbf{7,190}   &         7,465 & \textbf{7,177}   &         7,465 & \textbf{7,175}   &         7,465 & \textbf{7,174}   &         7,464 & \textbf{7,170} &         7,441 & \textbf{7,165} \\ 
C2\_6\_4      &         \textbf{7,306} & 7,323   &         7,180 & \textbf{7,018}   &         7,145 & \textbf{6,949}   &         7,132 & \textbf{6,947}   &         7,125 & \textbf{6,930}   &         7,123 & \textbf{6,928}   &         7,120 & \textbf{6,921} &         7,112 & \textbf{6,908} \\ 
C2\_6\_6      &         8,141 & \textbf{7,729}   &         7,924 & \textbf{7,530}   &         7,922 & \textbf{7,486}   &         7,914 & \textbf{7,486}   &         7,912 & \textbf{7,479}   &         7,906 & \textbf{7,479}   &         7,883 & \textbf{7,475} &         7,880 & \textbf{7,471} \\ 
\midrule 
R1\_6\_1      &         \textbf{22,345} & 22,453   &         22,070 & \textbf{22,049}   &         21,989 & \textbf{21,896}   &         21,987 & \textbf{21,856}   &         21,903 & \textbf{21,785}   &         21,813 & \textbf{21,745}   &         21,813 & \textbf{21,661} &         21,712 & \textbf{21,431} \\ 
R1\_6\_10      &         18,542 & \textbf{18,445}   &         18,305 & \textbf{18,298}   &         18,264 & \textbf{18,218}   &         18,181 & \textbf{18,077}   &         18,111 & \textbf{18,047}   &         18,064 & \textbf{18,041}   &         17,984 & \textbf{17,862} &         17,840 & \textbf{17,756} \\ 
R1\_6\_4      &         \textbf{16,420} & 16,709   &         \textbf{16,358} & 16,359   &         \textbf{16,271} & 16,292   &         16,265 & \textbf{16,220}   &         16,260 & \textbf{16,218}   &         \textbf{16,193} & 16,206   &         \textbf{16,101} & 16,167 &         \textbf{15,949} & 15,994 \\ 
R1\_6\_6      &         \textbf{18,825} & 18,859   &         \textbf{18,630} & 18,696   &         \textbf{18,564} & 18,632   &         \textbf{18,505} & 18,560   &         \textbf{18,456} & 18,492   &         \textbf{18,427} & 18,435   &         18,399 & \textbf{18,358} &         18,218 & \textbf{18,132} \\ 
\midrule 
R2\_6\_1      &         \textbf{15,812} & 15,866   &         15,618 & \textbf{15,540}   &         15,510 & \textbf{15,433}   &         15,455 & \textbf{15,369}   &         15,453 & \textbf{15,302}   &         15,438 & \textbf{15,287}   &         15,437 & \textbf{15,248} &         15,343 & \textbf{15,200} \\ 
R2\_6\_10      &         12,704 & \textbf{12,627}   &         \textbf{12,295} & 12,423   &         12,319 & \textbf{12,290}   &         12,238 & \textbf{12,196}   &         12,152 & \textbf{12,145}   &         12,139 & \textbf{12,118}   &         12,109 & \textbf{12,031} &         12,109 & \textbf{11,890} \\ 
R2\_6\_4      &         8,293 & \textbf{8,275}   &         \textbf{8,163} & 8,174   &         \textbf{8,111} & 8,130   &         8,117 & \textbf{8,090}   &         8,105 & \textbf{8,046}   &         8,087 & \textbf{8,015}   &         8,053 & \textbf{7,986} &         8,063 & \textbf{7,965} \\ 
R2\_6\_6      &         \textbf{12,692} & 12,810   &         \textbf{12,392} & 12,450   &         \textbf{12,383} & 12,406   &         12,336 & \textbf{12,294}   &         12,298 & \textbf{12,291}   &         \textbf{12,252} & 12,261   &         \textbf{12,193} & 12,220 &         12,116 & \textbf{12,061} \\ 
\midrule 
RC1\_6\_1      &         \textbf{17,887} & 17,982   &         \textbf{17,642} & 17,778   &         17,666 & \textbf{17,629}   &         17,578 & \textbf{17,509}   &         17,478 & \textbf{17,422}   &         17,469 & \textbf{17,350}   &         17,407 & \textbf{17,319} &         17,294 & \textbf{17,141} \\ 
RC1\_6\_10      &         \textbf{16,425} & 16,467   &         \textbf{16,171} & 16,341   &         \textbf{16,149} & 16,224   &         \textbf{16,073} & 16,153   &         \textbf{16,064} & 16,103   &         \textbf{16,044} & 16,086   &         \textbf{15,967} & 16,021 &         \textbf{15,865} & 15,924 \\ 
RC1\_6\_4      &         \textbf{15,383} & 15,471   &         \textbf{15,242} & 15,344   &         \textbf{15,224} & 15,240   &         \textbf{15,174} & 15,179   &         15,152 & \textbf{15,128}   &         \textbf{15,094} & 15,095   &         \textbf{14,993} & 15,031 &         14,963 & \textbf{14,928} \\ 
RC1\_6\_6      &         17,340 & \textbf{17,316}   &         17,082 & \textbf{17,008}   &         17,145 & \textbf{16,974}   &         17,046 & \textbf{16,872}   &         \textbf{16,868} & 16,872   &         \textbf{16,834} & 16,865   &         16,799 & \textbf{16,786} &         16,704 & \textbf{16,631} \\ 
\midrule 
RC2\_6\_1      &         \textbf{12,511} & 12,598   &         \textbf{12,350} & 12,360   &         12,265 & \textbf{12,226}   &         12,236 & \textbf{12,121}   &         12,230 & \textbf{12,072}   &         12,224 & \textbf{12,047}   &         12,201 & \textbf{12,027} &         12,191 & \textbf{12,011} \\ 
RC2\_6\_10      &         \textbf{9,361} & 9,540   &         \textbf{9,250} & 9,327   &         9,263 & \textbf{9,235}   &         9,209 & \textbf{9,194}   &         9,182 & \textbf{9,151}   &         9,180 & \textbf{9,140}   &         9,172 & \textbf{9,081} &         9,111 & \textbf{9,021} \\ 
RC2\_6\_4      &         \textbf{7,492} & 7,586   &         \textbf{7,308} & 7,417   &         7,281 & \textbf{7,278}   &         7,245 & \textbf{7,188}   &         7,217 & \textbf{7,138}   &         7,198 & \textbf{7,132}   &         7,178 & \textbf{7,100} &         7,165 & \textbf{6,995} \\ 
RC2\_6\_6      &         11,504 & \textbf{11,460}   &         \textbf{11,232} & 11,238   &         \textbf{11,194} & 11,195   &         11,150 & \textbf{11,076}   &         11,085 & \textbf{11,022}   &         11,068 & \textbf{11,006}   &         11,053 & \textbf{10,936} &         11,042 & \textbf{10,852} \\ 
\midrule 
\bottomrule
\end{tabular}

    \end{threeparttable}
\end{table}
\end{landscape}
\begin{landscape}
\begin{table}[ht]
    \centering
    \small
    \begin{threeparttable}
        \caption{\label{tab:dri_hgs_tw_800} Comparison of solution quality between DRI and HGS-TW for different runtimes - instance size: 800}
        \begin{tabular}{r|cc|cc|cc|cc|cc|cc|cc|cc}
\toprule
$\Theta$   &     \multicolumn{2}{c|}{15} &     \multicolumn{2}{c|}{30} &     \multicolumn{2}{c|}{45} &     \multicolumn{2}{c|}{60} &     \multicolumn{2}{c|}{75} &     \multicolumn{2}{c|}{90} &     \multicolumn{2}{c|}{120} &     \multicolumn{2}{c}{300} \\ 
{} &     DRI  & HGS &     DRI  & HGS &     DRI  & HGS &     DRI  & HGS &     DRI  & HGS &     DRI  & HGS &     DRI  & HGS &     DRI  & HGS \\ 
\midrule 
C1\_8\_1      &         \textbf{25,184} & \textbf{25,184}   &         \textbf{25,184} & \textbf{25,184}   &         \textbf{25,184} & \textbf{25,184}   &         \textbf{25,184} & \textbf{25,184}   &         \textbf{25,184} & \textbf{25,184}   &         \textbf{25,184} & \textbf{25,184}   &         \textbf{25,184} & \textbf{25,184} &         \textbf{25,184} & \textbf{25,184} \\ 
C1\_8\_10      &         \textbf{24,656} & 25,121   &         \textbf{24,427} & 24,914   &         \textbf{24,378} & 24,576   &         \textbf{24,349} & 24,576   &         \textbf{24,299} & 24,485   &         \textbf{24,295} & 24,423   &         \textbf{24,269} & 24,368 &         \textbf{24,209} & 24,271 \\ 
C1\_8\_4      &         \textbf{24,410} & 25,051   &         \textbf{24,264} & 24,586   &         \textbf{24,283} & 24,467   &         \textbf{24,170} & 24,389   &         \textbf{24,143} & 24,317   &         \textbf{24,094} & 24,287   &         \textbf{24,066} & 24,259 &         \textbf{23,978} & 24,024 \\ 
C1\_8\_6      &         25,163 & \textbf{25,160}   &         25,162 & \textbf{25,160}   &         25,162 & \textbf{25,160}   &         25,162 & \textbf{25,160}   &         25,162 & \textbf{25,160}   &         25,162 & \textbf{25,160}   &         25,162 & \textbf{25,160} &         25,162 & \textbf{25,160} \\ 
\midrule 
C2\_8\_1      &         \textbf{12,063} & 12,065   &         12,049 & \textbf{11,665}   &         12,049 & \textbf{11,662}   &         12,049 & \textbf{11,662}   &         12,049 & \textbf{11,662}   &         12,049 & \textbf{11,662}   &         12,049 & \textbf{11,662} &         12,049 & \textbf{11,662} \\ 
C2\_8\_10      &         \textbf{11,861} & 12,424   &         \textbf{11,421} & 11,688   &         11,378 & \textbf{11,235}   &         11,283 & \textbf{11,107}   &         11,232 & \textbf{11,052}   &         11,208 & \textbf{11,043}   &         11,188 & \textbf{11,023} &         11,160 & \textbf{11,004} \\ 
C2\_8\_4      &         \textbf{11,461} & 11,848   &         \textbf{11,083} & 11,290   &         \textbf{11,054} & 11,056   &         11,034 & \textbf{11,006}   &         11,011 & \textbf{10,934}   &         10,996 & \textbf{10,848}   &         10,985 & \textbf{10,762} &         10,936 & \textbf{10,720} \\ 
C2\_8\_6      &         \textbf{12,027} & 12,376   &         11,623 & \textbf{11,472}   &         11,591 & \textbf{11,377}   &         11,577 & \textbf{11,356}   &         11,544 & \textbf{11,356}   &         11,577 & \textbf{11,354}   &         11,573 & \textbf{11,349} &         11,573 & \textbf{11,347} \\ 
\midrule 
R1\_8\_1      &         \textbf{38,179} & 39,765   &         \textbf{37,689} & 38,715   &         \textbf{37,513} & 38,041   &         \textbf{37,484} & 37,819   &         \textbf{37,460} & 37,787   &         \textbf{37,416} & 37,752   &         \textbf{37,341} & 37,457 &         37,172 & \textbf{37,140} \\ 
R1\_8\_10      &         \textbf{32,545} & 32,886   &         \textbf{32,415} & 32,619   &         \textbf{32,140} & 32,406   &         \textbf{32,003} & 32,305   &         \textbf{31,877} & 32,171   &         \textbf{31,759} & 32,125   &         \textbf{31,658} & 32,085 &         \textbf{31,476} & 31,740 \\ 
R1\_8\_4      &         \textbf{29,182} & 29,365   &         \textbf{28,914} & 29,250   &         \textbf{28,908} & 29,047   &         \textbf{28,761} & 29,047   &         \textbf{28,719} & 28,982   &         \textbf{28,684} & 28,842   &         \textbf{28,599} & 28,726 &         28,182 & \textbf{28,169} \\ 
R1\_8\_6      &         \textbf{32,697} & 32,987   &         \textbf{32,453} & 32,789   &         \textbf{32,361} & 32,570   &         \textbf{32,309} & 32,466   &         \textbf{32,131} & 32,388   &         \textbf{32,103} & 32,388   &         \textbf{31,947} & 32,078 &         \textbf{31,604} & 31,840 \\ 
\midrule 
R2\_8\_1      &         \textbf{26,626} & 27,772   &         \textbf{26,038} & 26,560   &         \textbf{25,866} & 25,995   &         25,818 & \textbf{25,693}   &         25,783 & \textbf{25,554}   &         25,695 & \textbf{25,553}   &         25,550 & \textbf{25,452} &         25,403 & \textbf{25,152} \\ 
R2\_8\_10      &         21,753 & \textbf{21,629}   &         \textbf{21,006} & 21,141   &         \textbf{20,917} & 20,948   &         \textbf{20,790} & 20,886   &         \textbf{20,700} & 20,764   &         \textbf{20,582} & 20,728   &         20,616 & \textbf{20,596} &         20,503 & \textbf{20,119} \\ 
R2\_8\_4      &         \textbf{14,066} & 14,528   &         \textbf{13,873} & 14,161   &         \textbf{13,779} & 13,947   &         \textbf{13,707} & 13,840   &         \textbf{13,656} & 13,781   &         \textbf{13,569} & 13,741   &         \textbf{13,525} & 13,720 &         13,433 & \textbf{13,430} \\ 
R2\_8\_6      &         \textbf{21,442} & 21,854   &         \textbf{20,818} & 21,218   &         \textbf{20,794} & 21,118   &         \textbf{20,635} & 20,928   &         \textbf{20,591} & 20,698   &         \textbf{20,578} & 20,609   &         \textbf{20,494} & 20,519 &         20,175 & \textbf{20,171} \\ 
\midrule 
RC1\_8\_1      &         \textbf{31,275} & 31,837   &         \textbf{30,970} & 31,597   &         \textbf{30,907} & 31,437   &         \textbf{30,839} & 31,253   &         \textbf{30,768} & 30,983   &         \textbf{30,721} & 30,961   &         \textbf{30,667} & 30,867 &         \textbf{30,408} & 30,569 \\ 
RC1\_8\_10      &         \textbf{29,355} & 29,657   &         \textbf{29,121} & 29,325   &         \textbf{29,062} & 29,152   &         \textbf{29,022} & 29,130   &         \textbf{28,953} & 29,029   &         \textbf{28,890} & 29,008   &         \textbf{28,738} & 28,867 &         \textbf{28,518} & 28,577 \\ 
RC1\_8\_4      &         \textbf{27,685} & 27,959   &         \textbf{27,393} & 27,699   &         \textbf{27,432} & 27,532   &         \textbf{27,379} & 27,429   &         \textbf{27,282} & 27,360   &         \textbf{27,255} & 27,336   &         \textbf{27,097} & 27,182 &         26,983 & \textbf{26,951} \\ 
RC1\_8\_6      &         \textbf{30,531} & 30,876   &         \textbf{30,330} & 30,614   &         \textbf{30,060} & 30,457   &         \textbf{30,011} & 30,423   &         \textbf{29,909} & 30,213   &         \textbf{29,914} & 30,193   &         \textbf{29,782} & 30,035 &         \textbf{29,629} & 29,777 \\ 
\midrule 
RC2\_8\_1      &         \textbf{20,551} & 21,316   &         \textbf{19,946} & 20,127   &         \textbf{19,941} & 19,954   &         \textbf{19,785} & 19,817   &         19,704 & \textbf{19,607}   &         19,701 & \textbf{19,498}   &         19,667 & \textbf{19,435} &         19,617 & \textbf{19,318} \\ 
RC2\_8\_10      &         15,697 & \textbf{15,626}   &         15,094 & \textbf{15,057}   &         15,039 & \textbf{15,005}   &         15,098 & \textbf{14,827}   &         14,911 & \textbf{14,741}   &         14,911 & \textbf{14,666}   &         14,894 & \textbf{14,615} &         14,731 & \textbf{14,539} \\ 
RC2\_8\_4      &         \textbf{11,785} & 11,985   &         \textbf{11,518} & 11,704   &         \textbf{11,428} & 11,518   &         \textbf{11,399} & 11,442   &         11,369 & \textbf{11,341}   &         11,354 & \textbf{11,302}   &         11,343 & \textbf{11,231} &         11,248 & \textbf{11,100} \\ 
RC2\_8\_6      &         \textbf{18,524} & 19,154   &         \textbf{17,873} & 18,228   &         \textbf{17,824} & 17,880   &         17,709 & \textbf{17,648}   &         17,621 & \textbf{17,597}   &         17,611 & \textbf{17,518}   &         17,549 & \textbf{17,435} &         17,503 & \textbf{17,273} \\ 
\midrule 
\bottomrule
\end{tabular}

    \end{threeparttable}
\end{table}
\end{landscape}

\begin{landscape}
    \begin{table}[ht]
        \centering
        \small
        \caption{\label{tab:dri_hgs_tw_1000} Comparison of solution quality between DRI and HGS-TW for different runtimes - instance size: 1000}
        \begin{tabular}{r|cc|cc|cc|cc|cc|cc|cc|cc}
\toprule
$\Theta$   &     \multicolumn{2}{c|}{15} &     \multicolumn{2}{c|}{30} &     \multicolumn{2}{c|}{45} &     \multicolumn{2}{c|}{60} &     \multicolumn{2}{c|}{75} &     \multicolumn{2}{c|}{90} &     \multicolumn{2}{c|}{120} &     \multicolumn{2}{c}{300} \\ 
{} &     DRI  & HGS &     DRI  & HGS &     DRI  & HGS &     DRI  & HGS &     DRI  & HGS &     DRI  & HGS &     DRI  & HGS &     DRI  & HGS \\ 
\midrule 
C1\_10\_1      &         42,499 & \textbf{42,479}   &         42,499 & \textbf{42,479}   &         42,499 & \textbf{42,479}   &         42,499 & \textbf{42,479}   &         42,499 & \textbf{42,479}   &         42,499 & \textbf{42,479}   &         42,499 & \textbf{42,479} &         42,499 & \textbf{42,479} \\ 
C1\_10\_10      &         \textbf{41,440} & 42,629   &         \textbf{41,038} & 41,853   &         \textbf{40,996} & 41,596   &         \textbf{40,856} & 41,596   &         \textbf{40,846} & 41,435   &         \textbf{40,878} & 41,375   &         \textbf{40,660} & 41,234 &         \textbf{40,288} & 40,884 \\ 
C1\_10\_4      &         \textbf{40,721} & 42,189   &         \textbf{40,447} & 41,490   &         \textbf{40,423} & 41,387   &         \textbf{40,365} & 41,135   &         \textbf{40,160} & 41,019   &         \textbf{40,173} & 41,019   &         \textbf{40,045} & 41,019 &         \textbf{39,743} & 40,138 \\ 
C1\_10\_6      &         \textbf{42,492} & 42,499   &         42,491 & \textbf{42,473}   &         42,491 & \textbf{42,471}   &         42,491 & \textbf{42,471}   &         42,491 & \textbf{42,471}   &         42,491 & \textbf{42,471}   &         42,491 & \textbf{42,471} &         42,491 & \textbf{42,471} \\ 
\midrule 
C2\_10\_1      &         \textbf{17,837} & 19,065   &         17,440 & \textbf{17,404}   &         17,437 & \textbf{17,331}   &         17,437 & \textbf{16,879}   &         17,437 & \textbf{16,879}   &         17,437 & \textbf{16,879}   &         17,437 & \textbf{16,879} &         17,437 & \textbf{16,879} \\ 
C2\_10\_10      &         \textbf{16,835} & 19,317   &         \textbf{16,173} & 17,532   &         \textbf{16,079} & 16,700   &         \textbf{16,056} & 16,366   &         \textbf{16,012} & 16,067   &         15,996 & \textbf{15,990}   &         15,987 & \textbf{15,850} &         15,943 & \textbf{15,802} \\ 
C2\_10\_4      &         \textbf{16,879} & 18,076   &         \textbf{15,912} & 17,300   &         \textbf{15,901} & 16,727   &         \textbf{15,825} & 16,428   &         \textbf{15,688} & 15,904   &         \textbf{15,783} & 15,838   &         15,681 & \textbf{15,632} &         15,611 & \textbf{15,541} \\ 
C2\_10\_6      &         \textbf{17,390} & 19,038   &         \textbf{16,758} & 16,827   &         16,725 & \textbf{16,550}   &         16,695 & \textbf{16,398}   &         16,680 & \textbf{16,385}   &         16,668 & \textbf{16,365}   &         16,668 & \textbf{16,346} &         16,661 & \textbf{16,336} \\ 
\midrule 
R1\_10\_1      &         \textbf{57,588} & 291,558   &         \textbf{56,175} & 291,558   &         \textbf{55,729} & 291,558   &         \textbf{55,506} & 291,558   &         \textbf{55,393} & 56,086   &         \textbf{55,169} & 55,530   &         \textbf{55,052} & 55,227 &         \textbf{54,461} & 54,706 \\ 
R1\_10\_10      &         \textbf{50,181} & 50,389   &         \textbf{49,635} & 50,322   &         \textbf{49,793} & 49,939   &         \textbf{49,532} & 49,667   &         \textbf{49,242} & 49,406   &         \textbf{49,194} & 49,217   &         \textbf{48,914} & 49,160 &         \textbf{48,360} & 48,776 \\ 
R1\_10\_4      &         \textbf{44,620} & 45,312   &         \textbf{44,490} & 44,720   &         \textbf{44,434} & 44,548   &         \textbf{44,349} & 44,448   &         \textbf{44,291} & 44,297   &         \textbf{44,153} & 44,297   &         \textbf{43,786} & 44,123 &         \textbf{43,272} & 43,624 \\ 
R1\_10\_6      &         \textbf{49,909} & 50,374   &         \textbf{49,765} & 49,985   &         \textbf{49,496} & 49,827   &         \textbf{49,221} & 49,656   &         \textbf{49,393} & 49,603   &         \textbf{49,187} & 49,331   &         \textbf{48,968} & 49,176 &         \textbf{48,264} & 48,662 \\ 
\midrule 
R2\_10\_1      &         \textbf{39,652} & 41,491   &         \textbf{38,777} & 39,546   &         \textbf{38,420} & 38,719   &         \textbf{38,359} & 38,474   &         38,150 & \textbf{38,097}   &         \textbf{37,936} & 38,012   &         37,868 & \textbf{37,735} &         37,447 & \textbf{37,291} \\ 
R2\_10\_10      &         \textbf{32,690} & 33,478   &         \textbf{31,937} & 32,070   &         31,801 & \textbf{31,794}   &         \textbf{31,487} & 31,575   &         \textbf{31,328} & 31,377   &         31,262 & \textbf{31,182}   &         31,148 & \textbf{31,044} &         30,771 & \textbf{30,640} \\ 
R2\_10\_4      &         \textbf{19,561} & 20,050   &         \textbf{19,112} & 19,599   &         \textbf{19,129} & 19,275   &         \textbf{19,001} & 19,093   &         \textbf{18,675} & 19,093   &         \textbf{18,740} & 18,999   &         \textbf{18,566} & 18,784 &         18,372 & \textbf{18,347} \\ 
R2\_10\_6      &         \textbf{32,019} & 32,675   &         \textbf{31,335} & 32,070   &         \textbf{30,998} & 31,450   &         \textbf{30,867} & 31,172   &         \textbf{30,639} & 30,997   &         \textbf{30,619} & 30,753   &         \textbf{30,364} & 30,590 &         \textbf{30,113} & 30,306 \\ 
\midrule 
RC1\_10\_1      &         \textbf{48,414} & 48,767   &         \textbf{48,085} & 48,404   &         \textbf{47,659} & 48,053   &         \textbf{47,455} & 48,053   &         \textbf{47,579} & 47,876   &         \textbf{47,348} & 47,839   &         \textbf{47,345} & 47,483 &         \textbf{46,889} & 47,119 \\ 
RC1\_10\_10      &         \textbf{45,715} & 46,431   &         \textbf{45,401} & 45,829   &         \textbf{45,129} & 45,646   &         \textbf{45,062} & 45,454   &         \textbf{45,020} & 45,400   &         \textbf{44,826} & 45,251   &         \textbf{44,699} & 44,969 &         \textbf{44,296} & 44,480 \\ 
RC1\_10\_4      &         \textbf{43,278} & 43,517   &         \textbf{42,952} & 43,417   &         \textbf{42,857} & 43,319   &         \textbf{42,786} & 43,243   &         \textbf{42,705} & 43,117   &         \textbf{42,697} & 43,065   &         \textbf{42,501} & 42,979 &         \textbf{42,156} & 42,414 \\ 
RC1\_10\_6      &         \textbf{47,526} & 47,981   &         \textbf{46,856} & 47,312   &         \textbf{46,940} & 47,059   &         \textbf{46,654} & 46,816   &         46,658 & \textbf{46,603}   &         \textbf{46,408} & 46,603   &         \textbf{46,327} & 46,548 &         \textbf{45,720} & 46,048 \\ 
\midrule 
RC2\_10\_1      &         \textbf{30,590} & 34,143   &         \textbf{29,631} & 31,718   &         \textbf{29,376} & 30,520   &         \textbf{29,206} & 30,164   &         \textbf{29,158} & 29,856   &         \textbf{29,054} & 29,461   &         29,034 & \textbf{28,988} &         28,725 & \textbf{28,426} \\ 
RC2\_10\_10      &         \textbf{23,877} & 24,454   &         \textbf{23,086} & 23,574   &         22,980 & \textbf{22,942}   &         \textbf{22,884} & 22,933   &         22,765 & \textbf{22,704}   &         \textbf{22,668} & 22,675   &         \textbf{22,477} & 22,579 &         22,303 & \textbf{22,109} \\ 
RC2\_10\_4      &         \textbf{17,081} & 17,900   &         \textbf{16,665} & 17,500   &         \textbf{16,528} & 16,901   &         \textbf{16,500} & 16,714   &         \textbf{16,418} & 16,655   &         \textbf{16,362} & 16,543   &         \textbf{16,294} & 16,351 &         16,059 & \textbf{16,031} \\ 
RC2\_10\_6      &         \textbf{28,476} & 30,474   &         \textbf{27,305} & 29,121   &         \textbf{26,966} & 27,981   &         \textbf{26,976} & 27,465   &         \textbf{26,874} & 27,106   &         \textbf{26,779} & 26,991   &         \textbf{26,611} & 26,634 &         26,378 & \textbf{26,205} \\ 
\midrule 
\bottomrule
\end{tabular}

    \end{table}
\end{landscape}

\bibliographystyle{informs2014trsc} 
\bibliography{bibliography}  

\begin{thebibliography}{35}
\providecommand{\natexlab}[1]{#1}
\providecommand{\url}[1]{\texttt{#1}}
\providecommand{\urlprefix}{URL }

\bibitem[{Accorsi \protect\BIBand{} Vigo(2021)}]{Accorsi_a_21_filo}
Accorsi L, Vigo D (2021) A fast and scalable heuristic for the solution of
  large-scale capacitated vehicle routing problems. \emph{Transportation
  Science} 55(4):832--856.

\bibitem[{Anderberg(1973)}]{Anderberg_73_clust}
Anderberg MR (1973) \emph{Cluster Analysis for Applications} ({Academic Press},
  New York).

\bibitem[{Arnold et~al.(2019)Arnold, Gendreau, \protect\BIBand{}
  S{\"o}rensen}]{Arnold_19_v_large_scale}
Arnold F, Gendreau M, S{\"o}rensen K (2019) Efficiently solving very
  large-scale routing problems. \emph{Computers {\&} Operations Research}
  107:32--42.

\bibitem[{Arnold et~al.(2021)Arnold, Santana, S{\"o}rensen, \protect\BIBand{}
  Vidal}]{Arnold_a_21_PILS}
Arnold F, Santana {\'I}, S{\"o}rensen K, Vidal T (2021) {PILS: E}xploring
  high-order neighborhoods by pattern mining and injection. \emph{Pattern
  Recognition} 116:107957.

\bibitem[{Arnold \protect\BIBand{} S{\"o}rensen(2019)}]{Arnold_19_gls_pruning}
Arnold F, S{\"o}rensen K (2019) Knowledge-guided local search for the vehicle
  routing problem. \emph{Computers {\&} Operations Research} 105:32--46.

\bibitem[{Battarra et~al.(2014)Battarra, Erdo{\u{g}}an, \protect\BIBand{}
  Vigo}]{Battarra_14_cluvrp}
Battarra M, Erdo{\u{g}}an G, Vigo D (2014) Exact algorithms for the clustered
  vehicle routing problem. \emph{Operations Research} 62(1):58--71.

\bibitem[{Beek et~al.(2018)Beek, Raa, Dullaert, \protect\BIBand{}
  Vigo}]{Beek_a_18_pruning_ls}
Beek O, Raa B, Dullaert W, Vigo D (2018) An efficient implementation of a
  static move descriptor-based local search heuristic. \emph{Computers \&
  Operations Research} 94:1--10.

\bibitem[{Bent \protect\BIBand{}
  Van~Hentenryck(2010)}]{Bent_a_10_temp_decomp_vrp}
Bent R, Van~Hentenryck P (2010) Spatial, temporal, and hybrid decompositions
  for large-scale vehicle routing with time windows. Cohen D, ed.
  \emph{{Principles and Practice of Constraint Programming. }} (Springer,
  Berlin Heidelberg), 99--113.

\bibitem[{Bezdek(1981)}]{Bezdek_81_fcm}
Bezdek JC (1981) \emph{Pattern Recognition with Fuzzy Objective Function
  Algorithms} ({Springer US}, Boston).

\bibitem[{Br{\"a}ysy \protect\BIBand{} Gendreau(2005)}]{Braysy_05_VRPTW}
Br{\"a}ysy O, Gendreau M (2005) Vehicle routing problem with time windows,
  {Part I}: Route construction and local search algorithms.
  \emph{Transportation Science} 39(1):104--118.

\bibitem[{Clarke \protect\BIBand{} Wright(1964)}]{Clarke_a_64_sav}
Clarke G, Wright JW (1964) Scheduling of vehicles from a central depot to a
  number of delivery points. \emph{Operations Research} 12(4):568--581.

\bibitem[{Costa et~al.(2019)Costa, Contardo, \protect\BIBand{}
  Desaulniers}]{Costa_a_19_VRP_exact}
Costa L, Contardo C, Desaulniers G (2019) Exact branch-price-and-cut algorithms
  for vehicle routing. \emph{Transportation Science} 53(4):946--985.

\bibitem[{Dantzig \protect\BIBand{} Ramser(1959)}]{Dantzig_1959_vrp}
Dantzig GB, Ramser JH (1959) The truck dispatching problem. \emph{Management
  Science} 6(1):80--91.

\bibitem[{Ewbank et~al.(2016)Ewbank, Wanke, \protect\BIBand{}
  Hadi-Vencheh}]{Ewbank_16_fuzzy_clst_cvrp}
Ewbank H, Wanke P, Hadi-Vencheh A (2016) An unsupervised fuzzy clustering
  approach to the capacitated vehicle routing problem. \emph{Neural Computing
  and Applications} 27(4):857--867.

\bibitem[{Fisher \protect\BIBand{} Jaikumar(1981)}]{Fisher_81_gap_vrp}
Fisher ML, Jaikumar R (1981) A generalized assignment heuristic for vehicle
  routing. \emph{Networks} 11(2):109--124.

\bibitem[{Gehring \protect\BIBand{} Homberger(1999)}]{gehring_99_bm_vrptw}
Gehring H, Homberger J (1999) A parallel hybrid evolutionary metaheuristic for
  the vehicle routing problem with time windows. Miettinen K, M{\"a}kel{\"a} M,
  Toivanen J, eds. \emph{{Proc. EUROGEN99 }} (University of Jyväskylä,
  Finland, Finland), 57--64.

\bibitem[{Gillett \protect\BIBand{} Miller(1974)}]{Gillett_74_sweep}
Gillett BE, Miller LR (1974) A heuristic algorithm for the vehicle-dispatch
  problem. \emph{Operations Research} 22(2):340--349.

\bibitem[{Helsgaun(2000)}]{Helsgaun_00_effective}
Helsgaun K (2000) An effective implementation of the {L}in--{K}ernighan
  traveling salesman heuristic. \emph{European Journal of Operational Research}
  126(1):106--130.

\bibitem[{Jain(2010)}]{Jain_10_clst_survey}
Jain AK (2010) Data clustering: 50 years beyond k-means. \emph{Pattern
  Recognition Letters} 31(8):651--666.

\bibitem[{Kool et~al.(2022)Kool, Juninck, Roos, Cornelissen, Agterberg, van
  Hoorn, \protect\BIBand{} Visser}]{Kool_22_hgs_tw_impl}
Kool W, Juninck JO, Roos E, Cornelissen K, Agterberg P, van Hoorn J, Visser T
  (2022) Hybrid genetic search for the vehicle routing problem with time
  windows: {A} high-performance implementation. \emph{12th DIMACS
  Implementation Challenge Workshop}.

\bibitem[{Lin \protect\BIBand{} Kernighan(1973)}]{Lin_a_73_2opt}
Lin S, Kernighan BW (1973) An effective heuristic algorithm for the
  traveling-salesman problem. \emph{Operations Research} 21(2):498--516.

\bibitem[{Park \protect\BIBand{} Jun(2009)}]{Park_a_09_kmedoids}
Park HS, Jun CH (2009) A simple and fast algorithm for k-medoids clustering.
  \emph{Expert Systems with Applications} 36(2):3336--3341.

\bibitem[{Qi et~al.(2012)Qi, Lin, Li, \protect\BIBand{}
  Miao}]{Qi_a_12_st_clst_vrp}
Qi M, Lin WH, Li N, Miao L (2012) A spatiotemporal partitioning approach for
  large-scale vehicle routing problems with time windows. \emph{Transportation
  Research Part E: Logistics and Transportation Review} 48(1):248--257.

\bibitem[{Ropke \protect\BIBand{} Pisinger(2006)}]{Ropke_a_06_ALNS}
Ropke S, Pisinger D (2006) An adaptive large neighborhood search heuristic for
  the pickup and delivery problem with time windows. \emph{Transportation
  Science} 40(4):455--472.

\bibitem[{{S}ahin \protect\BIBand{} Yaman(2022)}]{Sahin_22_exact_VRPTW}
{S}ahin MK, Yaman H (2022) A branch and price algorithm for the heterogeneous
  fleet multi-depot multi-trip vehicle routing problem with time windows.
  \emph{Transportation Science} 56(6):1636--1657.

\bibitem[{Santini et~al.(2023)Santini, Schneider, Vidal, \protect\BIBand{}
  Vigo}]{Santini_a_23_decomp_HGS}
Santini A, Schneider M, Vidal T, Vigo D (2023) Decomposition strategies for
  vehicle routing heuristics. \emph{INFORMS Journal on Computing}
  35(3):543--559.

\bibitem[{Schneider et~al.(2015)Schneider, Stenger, Schwahn, \protect\BIBand{}
  Vigo}]{Schneider_15_tbra_tw}
Schneider M, Stenger A, Schwahn F, Vigo D (2015) Territory-based vehicle
  routing in the presence of time-window constraints. \emph{Transportation
  Science} 49(4):732--751.

\bibitem[{Solomon(1987)}]{Solomon_87_vrptw_bm}
Solomon MM (1987) Algorithms for the vehicle routing and scheduling problems
  with time window constraints. \emph{Operations Research} 35(2):254--265.

\bibitem[{Toth \protect\BIBand{} Vigo(2003)}]{Toth_a_03_pruning}
Toth P, Vigo D (2003) The granular tabu search and its application to the
  vehicle routing problem. \emph{INFORMS Journal on Computing} 15(4):333--346.

\bibitem[{Toth \protect\BIBand{} Vigo(2014)}]{Toth_2014_vrp}
Toth P, Vigo D (2014) \emph{Vehicle Routing: Problems, Methods, and
  Applications}{, 2nd ed.} ({Society for Industrial and Applied Mathematics},
  Philadelphia).

\bibitem[{Vidal et~al.(2012)Vidal, Crainic, Gendreau, Lahrichi,
  \protect\BIBand{} Rei}]{Vidal_12_HGS_org}
Vidal T, Crainic TG, Gendreau M, Lahrichi N, Rei W (2012) A hybrid genetic
  algorithm for multidepot and periodic vehicle routing problems.
  \emph{Operations Research} 60(3):611--624.

\bibitem[{Vidal et~al.(2013)Vidal, Crainic, Gendreau, \protect\BIBand{}
  Prins}]{Vidal_13_HGS-TW}
Vidal T, Crainic TG, Gendreau M, Prins C (2013) A hybrid genetic algorithm with
  adaptive diversity management for a large class of vehicle routing problems
  with time-windows. \emph{Computers {\&} Operations Research} 40(1):475--489.

\bibitem[{Vidal et~al.(2020)Vidal, Laporte, \protect\BIBand{}
  Matl}]{Vidal_20_vrp_variants}
Vidal T, Laporte G, Matl P (2020) A concise guide to existing and emerging
  vehicle routing problem variants. \emph{European Journal of Operational
  Research} 286(2):401--416.

\bibitem[{Wong \protect\BIBand{} Beasley(1984)}]{Wong_84_del_area}
Wong KF, Beasley JE (1984) Vehicle routing using fixed delivery areas.
  \emph{Omega} 12(6):591--600.

\bibitem[{Y{\"u}cenur \protect\BIBand{}
  Demirel(2011)}]{Yucenur_11_clst_ga_mdvrp}
Y{\"u}cenur GN, Demirel N{\c{C}} (2011) A new geometric shape-based genetic
  clustering algorithm for the multi-depot vehicle routing problem.
  \emph{Expert Systems with Applications} 38(9):11859--11865.

\end{thebibliography}






\end{document}